\documentclass[lettersize,journal]{IEEEtran}
\usepackage{amsmath,amsfonts}
\usepackage[ruled,vlined]{algorithm2e}
\usepackage{array}
\usepackage[caption=false,font=normalsize,labelfont=sf,textfont=sf]{subfig}
\usepackage{textcomp}
\usepackage{stfloats}
\usepackage{url}
\usepackage{verbatim}
\usepackage{graphicx}
\usepackage{cite}
\usepackage{amssymb}
\usepackage{lipsum}
\usepackage{tikz}
\usepackage{mathrsfs}
\usepackage{mathtools}
\usepackage[ruled,vlined]{algorithm2e}

\definecolor{navyblue}{rgb}{0.0, 0.0, 0.5}
\definecolor{indigo}{rgb}{0.29, 0.0, 0.51}
\definecolor{linkpurple}{rgb}{0.58,0.2,0.82}

\usepackage{hyperref}
\hypersetup{
    colorlinks = true,
    urlcolor   = blue,
    citecolor  = navyblue,
}

\newcommand{\RomanNumeralCaps}[1]
\linenumbers

\DeclarePairedDelimiter{\abs}{\lvert}{\rvert}
\DeclarePairedDelimiter{\norm}{\big\lVert}{\big\rVert}

\DeclarePairedDelimiter{\binner}{\big\langle}{\big\rangle}

\DeclarePairedDelimiter{\nint}\lfloor\rceil

\newcommand{\invcgi}{{\mathcal{C}^{-1}_{\bm{g}_i}}}

\newcommand{\invcgo}{{\mathcal{C}^{-1}_{\bm{g}_o}}}

\newcommand{\cov}{\mathcal{C}}
\newcommand{\R}{\mathbb{R}}

\newcommand{\zetaext}{{\mathring{\zeta}}}

\newcommand{\sdist}{{\phi_\pm}}

\newcommand*\mean[1]{\bar{#1}}

\newcommand\mapsfrom{\mathrel{\reflectbox{\ensuremath{\mapsto}}}}

\let\bm\boldsymbol

\newcommand{\rev}[1]{\textcolor{black}{#1}}
\newcommand{\revv}[1]{\textcolor{black}{#1}}

\newcommand{\revall}[1]{\textcolor{black}{#1}}

\begin{document}

\title{Physics-informed compressed sensing for PC-MRI: an inverse Navier--Stokes problem}

\author{Alexandros Kontogiannis, Matthew P. Juniper
\thanks{Engineering Department, University of Cambridge, Trumpington Street, Cambridge CB2
1PZ, UK}
}



\maketitle

\begin{abstract}
We formulate a physics-informed compressed sensing (PICS) method for the reconstruction of velocity fields from noisy and sparse phase-contrast magnetic resonance signals. The method solves an inverse Navier--Stokes boundary value problem, which permits us to jointly reconstruct and segment the velocity field, and at the same time infer hidden quantities such as the hydrodynamic pressure and the wall shear stress. Using a Bayesian framework, we regularize the problem by introducing \textit{a priori} information about the unknown parameters in the form of Gaussian random fields. This prior information is updated using the Navier--Stokes problem, an energy-based segmentation functional, and by requiring that the reconstruction is consistent with the $k$-space signals. We create an algorithm that solves this inverse problem, and test it for noisy and sparse $k$-space signals of the flow through a converging nozzle. We find that the method is capable of reconstructing and segmenting the velocity fields from sparsely-sampled (15\% $k$-space coverage), low ($\sim$$10$) signal-to-noise ratio (SNR) signals, and that the reconstructed velocity field compares well with that derived from fully-sampled (100\% $k$-space coverage) high ($>$$40$) SNR signals of the same flow. 
\end{abstract}


\section{Introduction}
\label{sec:intro}

\IEEEPARstart{I}{n} phase-contrast magnetic resonance imaging one seeks to reconstruct complex images (${w}$) whose magnitude is proportional to the nuclear spin density ($\rho$) and whose phase difference ($\Delta \varphi$) is proportional to a flow velocity component ($u_i$). For fully-sampled $k$-space signals (${s}^\bullet$) there is a one-to-one correspondence between the $k$-space and the physical (complex) space, which is given by a Fourier transform ($\mathcal{F}$) such that $s^\bullet = \mathcal{F} w$. Then, for sufficiently high ($ > 3$) signal-to-noise ratios (SNR), the noise in the magnitude and the phase of the complex image can be assumed to be white and Gaussian \cite{Gudbjartsson1995}. In this case, it is reasonable to directly reconstruct (denoise) the complex image in physical space, using either general purpose image denoising algorithms \cite{Chang2000,Pascal2012}, or physics-informed algorithms \cite{Fatouraee2003,Ong2015,Mura2016,Koltukluoglu2018,Funke2019,Kontogiannis2021}. For sparsely-sampled $k$-space signals (${s}^\star$) there is no longer a one-to-one correspondence between $k$-space and physical space. A naive solution to this problem is to zero-fill the subsampled $k$-space signal and then perform an inverse Fourier transform. This approach, known as the `zero-filling solution', reconstructs a complex image that is corrupted by artefacts and interference noise. Incoherent subsampling can eliminate artefacts, but interference noise remains a problem. In this case, it is better to reconstruct the complex image using compressed sensing (CS) \cite{Candes2006,Donoho2006,Lustig2007}.

\begin{figure}
\centering\hfill
\includegraphics[width=0.95\linewidth]{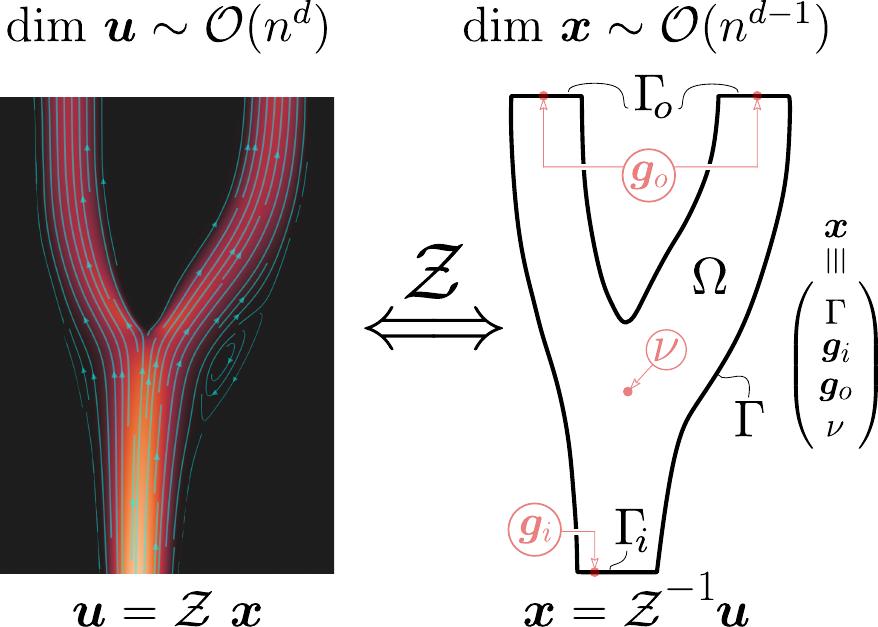}
\caption{\rev{A $d$-dimensional velocity field ($\bm{u}$) that can be described by a Navier--Stokes problem ($\mathcal{Z}$) has an underlying ($d-$1)-dimensional structure {($\bm{x}=\mathcal{Z}^{-1}\bm{u}$)}, which is the parameter vector containing the shape of the object ($\Gamma$), the boundary conditions $(\bm{g}_i,\bm{g}_o)$, and the kinematic viscosity ($\nu$). Images of $n^d$ voxels depicting the $d$ velocity components can be compressed/decompressed by solving an inverse/forward N--S problem.}}
\label{fig:sparsity_extended}
\end{figure}

Compressed sensing relies on the assumption that the complex image, or its constituent parts (e.g. the magnitude and the phase), \rev{have a sparse representation in a transform domain, and that the transform domain is considered to be known \textit{a priori}}. The reconstructed complex image can then be recovered by solving a (generally) nonlinear optimization problem of the form \cite{Benning2014}
\rev{$$ \min_{w}\ \norm{Tw}_{\ell^1} \quad \text{subject to}\quad \norm{s^\star-\mathcal{P}\mathcal{F}w}_{\ell^2} < \sqrt{N_s} \varepsilon\quad,$$}
where $T$ is a sparsity-inducing transformation, $\mathcal{P}$ is the sparse sampling pattern, $N_s$ is the total number of sampled $k$-space points, and $\varepsilon \propto \sigma$ is a user-selected value proportional to the standard deviation of noise ($\sigma$). In the above example, the functional to be minimized acts as a regularizer while the constraint enforces consistency between the reconstruction and the measurement. It is also possible to decompose the complex image into real and imaginary \cite{Holland2010,Roberts2013}, or into magnitude and phase components \cite{Zhao2012}\cite[Chapter~9.3]{Reci2019}, and reconstruct them separately. With this decomposition, different regularizers can be used for each component. This is particularly important for PC-MRI because suitable regularizers for the reconstructed magnitude, e.g. total variation (TV), may produce large errors or artefacts (e.g. staircasing) in the reconstructed phase. This is because the magnitude image often has high-contrast features resembling a piecewise-constant function, which is sparse in the TV-transform domain \cite{Lustig2007,Benning2014}, while the phase usually has smooth features since it encodes velocity information, and thus is no longer sparse in the same domain. A suitable transform for the phases, and for piecewise-smooth functions in general, is second-order total generalized variation (TGV$^2$) \cite{Bredies2010,Benning2014}. Complications arise due to the fact that the phase of the complex image alone does not represent velocity, but the phase difference does. Therefore, if there is available \mbox{\textit{a priori}} information on the velocity, the regularization should be imposed on the phase difference rather than on the individual phases \cite[Chapter~9.3]{Reci2019}\cite{Corona2021,Kollmeier2022}.

Even though generic CS methods perform very well in magnitude reconstruction, accurate velocity reconstruction from subsampled $k$-space signals remains a challenge. It has been suggested \cite[Chapter~11.1]{Reci2019} that even sparser PC-MRI signals could be reconstructed if a regularization method based on the Navier--Stokes (N--S) {equations} is used. However, the effect of the boundary conditions should be equally important in order to capture the velocity profiles in both the lumen and the near-wall region, and to estimate the wall-shear stress\footnote{Wall shear stress is often sought-after in fluid mechanics applications.} with greater confidence. Based on these observations, we believe that a general and accurate way of injecting \textit{a priori} knowledge for velocity (phase difference) regularization is in the form of a N--S \textit{boundary value problem}, and not in the form of the N--S \textit{equations} alone. 

\revv{Several physics-informed velocity regularization methods have been proposed in the past \cite{Fatouraee2003,Roberts2013,Ong2015,Mura2016,Bakhshinejad2017,Koltukluoglu2018,Funke2019,Toger2020}, but none of them exploits the full structure of a Bayesian inverse Navier--Stokes problem in which the domain boundary ($\partial\Omega$), the boundary conditions ($\bm{g}_i,\bm{g}_o$), and the kinematic viscosity ($\nu$) are all considered unknown. A similar approach is discussed in \cite{Kontogiannis2021}, but the method applies only to fully-sampled PC-MRI signals and unwrapped velocity images.}

Solving an inverse N--S problem ($\mathcal{Z}$) amounts to finding its unknown parameters ($\bm{x}$), which produce a modelled velocity ($\bm{u} \equiv \mathcal{Z}\bm{x}$) that approximates the measured velocity ($\bm{u}^\star \propto \Delta\varphi$) in an appropriate norm. In this way, not only do we obtain a regularized, noiseless ($\text{SNR}=\infty$) velocity field, but we also infer hidden flow-related quantities such as the hydrodynamic pressure ($p$) and the wall shear stresses \cite{Sotelo2016,Zhang2020}, which cannot be measured using conventional MRI or PC-MRI methods, but which naturally arise from the N--S problem. It is also important to note that, because we jointly reconstruct and segment the velocity field, geometric errors are minimized \cite{Berg2014,Id2019}. 

Our approach to physics-informed CS extends the standard notion of sparsity used in conventional CS methods to a more general notion of a structure \cite{Duarte2011}, which is dictated by the \mbox{N--S} problem. Instead of enforcing sparsity during the nonlinear optimization process, we recover a sparse (hidden) structure of the velocity field by enforcing the Navier--Stokes problem as a constraint. This is better explained in figure \ref{fig:sparsity_extended}, which shows that the velocity exhibits an underlying low-dimensional structure that is encoded in the unknown parameters $\bm{x}$. Based on this, a velocity field can be compressed/decompressed by solving an inverse/forward Navier--Stokes problem. In contrast to our approach, a conventional sparsity-promoting and physics-informed velocity regularization method is described in \cite{Bright2013}, but it relies on a pre-existing database (library) containing the dominant eigenmodes of a Navier--Stokes problem that is defined in a pre-set geometry, with pre-set inlet boundary condition. 

\revall{Unlike pure machine learning and library-based algorithms \cite{Ferdian2020,Rutkowski2021}, which learn to recognize features in training data (often generated using computational fluid dynamics), our method encapsulates fluid mechanics knowledge in the form of the N--S problem. It can therefore reconstruct flows that it has not yet seen and extrapolate to new flow conditions, enabling patient-specific cardiovascular modelling. While neural networks (NNs) have revolutionized the field of computer vision, some fundamental problems still need to be addressed (e.g. \mbox{AI-generated} hallucinations, and existence of computational algorithms) \cite{9420272,doi:10.1073/pnas.2107151119}. Our method, on the other hand, is formulated in a variational framework and is therefore amenable to mathematical analysis. In the future this could provide a reference for the formulation of a rigorous \mbox{NN-based} algorithm that approximates the N--S solution and other partial differential equations using learned operators \cite{Bungert_2020,doi:10.1137/20M1338460}. To the best of our knowledge, there is currently neither a NN that can solve the \mbox{N--S} problem more efficiently than computational fluid dynamics methods, nor a NN that is proven to approximate the N--S boundary value \mbox{problem operator}.}

In this paper, we build upon the work of \cite{Kontogiannis2021} in order to formulate a physics-informed compressed sensing (PICS) method for the joint reconstruction and segmentation of velocity fields from sparse and noisy $k$-space signals. We provide an algorithm that solves the reconstruction problem and demonstrate it on \mbox{$k$-space} signals of a steady axisymmetric flow. \rev{Because the acquired signals were originally fully-sampled, we sparsify them using two different sparse sampling patterns and study the effect of sampling density on the velocity reconstruction error.} In section \ref{sec:pics_main} we formulate the physics-informed compressed sensing method, and an algorithm that implements it. In section \ref{sec:rec_results_main} we test the method on sparse and noisy $k$-space signals. 

\section{Physics-informed compressed sensing as an inverse Navier--Stokes problem}
\label{sec:pics_main}

In what follows, $L^2(\Omega)$ denotes the space of square-integrable functions in $\Omega$, with inner product $\binner{\cdot,\cdot}$ and norm $\norm{\cdot}_{L^2(\Omega)}$, and $H^k(\Omega)$ denotes the space of square-integrable functions with $k$ square-integrable derivatives. For a given covariance operator, $\mathcal{C}$, we also define the covariance-weighted $L^2(\Omega)$ spaces, endowed with the inner product ${\binner{\cdot,\cdot}_{\mathcal{C}(\Omega)} := \binner{\mathcal{C}^{-1/2}\cdot,\mathcal{C}^{-1/2}\cdot}}_{\Omega}$, which generates the norm $\norm{\cdot}_{\mathcal{C}(\Omega)}$. The Euclidean norm in the space of real numbers $\R^n$ is denoted by $\abs{\cdot}_{\R^n}$, and the measure (volume) of the domain $\Omega$ by $\abs{\Omega}$. The first variation of a functional $\mathscr{J}:L^2\to\mathbb{R}$ with respect to an unknown $z \in L^2$ is defined by
\begin{equation}
\delta_{z}\mathscr{J}\equiv\frac{d}{d\tau}\mathscr{J}({z}+\tau{z}',\dots)\Big\vert_{\tau=0}=\binner{D_z\mathscr{J},z'}\quad,
\label{eq:first_variation}
\end{equation}
where $\tau \in \mathbb{R}$, $z' \in L^2$ is an allowed perturbation of $z$, and $D_z\mathscr{J}$ is the generalized gradient. If $z$ is defined on a covariance-weighted $L^2$ space, we furthermore define the steepest ascent direction $\widehat{D}_z\mathscr{J}$ such that  
\begin{equation}
\binner{\widehat{D}_z\mathscr{J},z'}_\mathcal{C} = \binner{\mathcal{C}^{-1}\widehat{D}_z\mathscr{J},z'} = \binner{{D}_z\mathscr{J},z'}\quad,
\label{eq:sadirection}
\end{equation}
therefore $\widehat{D}_z\mathscr{J} = \mathcal{C}{D}_z\mathscr{J}$. When the covariance takes the form $\mathcal{C} = \sigma^2\mathrm{I}$, where $\sigma^2 \in \mathbb{R}$ is the variance and $\mathrm{I}$ is the identity operator, we write $\sigma^{-2}\norm{\cdot}_{L^2}$ instead of $\norm{\cdot}_\mathcal{C}$ for simplicity. \rev{We use the superscript $(\cdot)^\dagger$ to denote the adjoint of an operator,} $(\cdot)^\star$ to denote a measurement, $(\cdot)^\circ$ to denote a reconstruction, and $(\cdot)^\bullet$ to denote the ground truth. Note that, for the velocity, $\bm{u}^\star$ denotes the velocity obtained from the phase differences, and $\bm{u}$ denotes the velocity obtained from the Navier--Stokes problem. 

\subsection{Phase-contrast magnetic resonance imaging}
Phase-contrast magnetic resonance imaging can measure \mbox{$d$-dimensional} velocity fields $\bm{u}$ inside or around an object $\Omega$, by requiring a minimum of $d$ sets of $k$-space signals $\{\bm{s}\}_{i=1}^d$, one for each velocity component. The velocity component $u_i$ can be recovered from the signal set $\bm{s}_i=\{s\}_{j=1}^{4}$ by computing the phase difference
\begin{equation}
u_i = c_i(\varphi_1 - \varphi_2 - \varphi_3 + \varphi_4) \equiv c_i (\Delta \varphi)_i \quad,
\label{eq:phase_diff}
\end{equation}
where $c_i$ is a known constant that depends on the gyromagnetic ratio of hydrogen and the gradient pulse properties, and $\varphi_j \equiv \textrm{arg}\big(\mathcal{F}^{-1} s_j\big)$ when $s_j$ is a fully-sampled $k$-space signal. Note that the last two phases in the above phase difference correspond to zero-flow experiments in order to remove any phase
shift contributions that are not caused by the flow.

For sparsely-sampled $k$-space signals we define the sparse sampling operator $\mathcal{P}: I \to I_s$, which projects from the full image space $I \subset \mathbb{C}^d$ to the sparse image space $I_s \subset I$. Then, if $s_j$ is sparse, the corresponding complex image is given by
\begin{equation}
w_j \equiv \rho_j e^{i\varphi_j} = \mathcal{F}^{-1}\mathcal{P}^{-1}s_j \quad.
\label{eq:complex_image}
\end{equation}
Since the backprojection $\mathcal{P}^{-1}$ is ill-posed, we define $\mathcal{P}^{-1}s_j$ as the zero-filled $k$-space signal in $I$. We also obtain the zero-filling solution for the velocity by using equation \eqref{eq:phase_diff} with $\varphi_j = \textrm{arg}(w_j)$. As was mentioned in section \ref{sec:intro}, the zero-filling velocity solution is corrupted by interference noise and artefacts that strongly depend on the sparse sampling pattern $\mathcal{P}$. We therefore need to reconstruct the complex images $w_j$ using a different approach.

\subsection{Physics-informed compressed sensing (PICS) formulation}

\subsubsection{Phase regularization}
\label{subsec:phase_reg}
\textit{A priori} knowledge of the velocity, i.e. the phase difference, comes in the form of a Navier--Stokes boundary value problem. Assuming steady, incompressible flow and a Newtonian fluid, the fluid dynamics are governed by the following problem (see figure \ref{fig:sparsity_extended})
\begin{equation}
\left\{\begin{alignedat}{2}
    \bm{u}\bm{\cdot}\nabla\bm{u}-\nu{\Delta} \bm{u}  + \nabla p &= \bm{0} \quad &&\textrm{in}\quad \Omega \\
    \nabla \bm{\cdot} \bm{u} &= 0 \quad &&\textrm{in}\quad \Omega\\
    \bm{u} &= \bm{0} \quad &&\textrm{on}\quad \Gamma \\
   \bm{u} &= \bm{g}_i \quad &&\textrm{on}\quad \Gamma_i\\
   -\nu\partial_{\bm{\nu}}\bm{u}+p\bm{\nu} &= \bm{g}_o \quad &&\textrm{on}\quad \Gamma_o
  \end{alignedat}\right.\quad,
  \label{eq:navierstokes_bvp}
\end{equation}
where $\bm{u}$ is the velocity, $p\mapsfrom p/\rho$ is the reduced hydrodynamic pressure\footnote{From now on we refer to $p$ simply as the pressure.}, $\rho$ is the density, $\nu$ is the kinematic viscosity, $\bm{g}_i$ is the Dirichlet boundary condition at the inlet $\Gamma_i$, $\bm{g}_o$ is the natural boundary condition at the outlet $\Gamma_o$, $\bm{\nu}$ is the unit normal vector on $\partial\Omega$, and ${\partial_{\bm{\nu}}\equiv\bm{\nu}\bm{\cdot}\nabla}$ is the normal derivative. \rev{A zero-velocity (no-slip) boundary condition is imposed on the walls $\Gamma$ of the object $\Omega$.} We furthermore define the projection operator $\mathcal{S}: M \to I$ that projects from the model space $M$ to the image space $I$. The model space $M$ is a fixed space that always contains the deformable domain $\Omega$ ($\Omega \subset M$), such that $\Gamma_i,\Gamma_o \subset \partial M$, and
\begin{gather}
\partial M \cap \textrm{Int}\ \Gamma = \varnothing,\ \partial M \cap \Gamma_i = \partial M_i,\ \partial M \cap \Gamma_o = \partial M_o,
\end{gather}
where $\textrm{Int}$ denotes the interior of a set, i.e. $\textrm{Int}\ \Gamma \equiv \big(\Gamma-\partial\Gamma\big)$. The domain $\Omega$ does not have to be path-connected. In general, it can be defined as the union of disjoint, path-connected sets $\Omega_j$, i.e. $\Omega = \Omega_1 \cup \Omega_2 \cup \dots$, provided that every $\Omega_j$ includes at least one inlet and one outlet. 

The discrepancy between the measured ($\bm{u}^\star$) and the modelled ($\bm{u}$) velocity fields is given by
\begin{equation}
\mathscr{U}(\bm{\varphi},\bm{u}) \equiv \frac{1}{2}\sum^{d}_{i=1}\norm{{u}^\star_i - \mathcal{S}u_i}^2_{\mathcal{C}_{u_i}(I_w)}\quad,
\label{eq:velocity_phase_diff_discrepancy}
\end{equation}
where $\bm{\varphi}$ contains the phases needed to compute $\bm{u}^\star = \bm{u}^\star(\bm{\varphi})$ using \eqref{eq:phase_diff}, $I_w \subseteq I$ is a user-selected area (window) of interest, and $\mathcal{C}_{u_i}$ is the covariance operator for the velocity discrepancy of the $i$-th component. In \cite{Kontogiannis2021} we had assumed a diagonal covariance operator for the velocity discrepancy because the noise in the phase images can be assumed to be white and Gaussian for fully-sampled $k$-space signals with $\text{SNR} > 3$ \cite{Gudbjartsson1995}. Here, we model the interference (correlated) noise using an exponential covariance operator such that
\begin{equation}
\mathcal{C}_{u_i} g  \equiv \sigma^2_{u_i}\big(C*g\big)\quad \text{for any}\quad g \in L^2(I)\quad,
\label{eq:exp_cov_op}
\end{equation}
where $\sigma_{u_i}$ is the standard deviation of noise in the $i$-th velocity component, `$*$' denotes convolution, and $C$ is the exponential kernel
\begin{equation}
C(r) = \frac{1}{d!V_d(\ell)}\ e^{-\abs{r}/\ell} \quad \text{for any}\quad r \in \mathbb{R}^d\quad,
\label{eq:exp_cov_kernel}
\end{equation}
where $d$ is the dimension of the velocity field $\bm{u}$, $\ell$ is the characteristic length, $V_d(\ell) = \pi^{d/2}\ell^d/\Gamma(d/2+1)$ is the volume of the $d$-dimensional Euclidean ball of radius $\ell$, and $\Gamma$ is the gamma function. Adopting a Bayesian inference setting similar to \cite{Kontogiannis2021}, we introduce a $2\pi$-periodic prior for the phase \cite{Zhao2012}. The combined phase regularization functional is then given by 
\begin{align}
\mathscr{R}_\varphi(\bm{\varphi},\bm{u}) \equiv \mathscr{U} + \frac{1}{2}\sum^{4d}_{j=1}\norm{e^{i\varphi_j}-e^{i\bar{\varphi}_j}}^2_{\mathcal{C}_{\varphi_j}(I)}\quad,
\end{align}
where $\bar{\varphi}_j$ is the prior assumption of $\varphi_j$, $\mathcal{C}_{\varphi_j} = \widetilde{\sigma}_{{\varphi_j}}^2 (C * \cdot)$, $\widetilde{\sigma}_{{\varphi_j}}=\xi_\varphi \sigma_{{\varphi_j}}$, and $\xi_\varphi \in \R$ is a user-selected parameter that determines the confidence in the prior assumption.\\

\subsubsection{Navier--Stokes parameters regularization}
Additional regularization is required to penalize improbable solutions of the Navier--Stokes parameters $\bm{x}$. The parameters $\bm{x}$ consist of the shape of the domain $\partial\Omega$, the Dirichlet boundary condition at the inlet(s) $\bm{g}_i$, the natural boundary condition at the outlet(s) $\bm{g}_o$, and the kinematic viscosity $\nu$. Note that we identify the object $\Omega$ and its boundary with a signed distance function $\sdist \in L^2(M)$ such that \cite[Section~2.4]{Kontogiannis2021}
\begin{gather*}
\Omega =\big\{y \in \Omega:\ \sdist(y) < 0 \big\} \ ,\ \partial\Omega = \big\{y \in \Omega:\ \sdist(y) = 0 \big\}.
\label{eq:sdf_of_domain}
\end{gather*}
The regularization functional for $\bm{x}$ is given by \cite[Section~2.2]{Kontogiannis2021}
\begin{align}
\mathscr{R}_x(\bm{x}) = &\phantom{+}\frac{1}{2\sigma_\sdist^2}\norm{\mean{\phi}_\pm-\sdist}^2_{L^2(M)}+\frac{1}{2}\norm{\bm{g}_i-\mean{\bm{g}}_i}^2_{\cov_{\bm{g}_i}(\partial M_i)}\nonumber\\ &+\frac{1}{2}\norm{\bm{g}_o-\mean{\bm{g}}_o}^2_{\cov_{\bm{g}_o}(\partial M_o)} + \frac{1}{2\sigma_\nu^2}\big|\nu - \mean{\nu}\big|^2_{\mathbb{R}}\quad,
\end{align}
where $(\bar{\cdot})$ is a prior assumption, $\sigma_{(\cdot)}$ is the standard deviation of the prior, and $\cov_{\bm{g}_i},\cov_{\bm{g}_o}$ are covariance operators that are based on the exponential covariance kernel \eqref{eq:exp_cov_kernel}, but with additional boundary conditions, which are described in \cite{Kontogiannis2021}.
\\

\subsubsection{Magnitude segmentation}
To exploit information about $\partial\Omega$ from the nuclear spin density (magnitude) images we use an energy-based segmentation functional $\mathscr{S}$, which assumes that the image consists of two regions with approximately uniform magnitudes (e.g. stars in a night sky) \cite{Chan2001,Getreuer2012}. The functional is given by
\begin{align}
\mathscr{S}(\bm{\rho},\sdist,\alpha,\beta) &\equiv \frac{1}{8d}\sum^{4d}_{j=1}\frac{1}{\sigma^2_{\rho_j}}\bigg(\norm{\big(\rho_j-\alpha\big)\mathcal{S}\mathcal{H}(\sdist)}^2_{L^2(I_w)}\nonumber\\ 
+& \norm{\big(\rho_j-\beta\big)\big(\mathcal{S}\mathcal{H}(\sdist)-1\big)}^2_{L^2(I_w)} \bigg)\quad,\label{eq:segm_functional}
\end{align}
where the mean value of the average magnitude inside $\Omega$ is $\alpha \in \mathbb{R}$ and outside $\Omega$ is $\beta \in \mathbb{R}$. In addition, $\sigma_{\rho_j}$ is the standard deviation of noise in the magnitude image, and $\mathcal{H}$ denotes the (modified) Heaviside function such that
\begin{equation}
\mathcal{H}(t) = 1 \quad \text{if}\quad t < 0 \quad \text{else}\quad 0\quad.
\end{equation}
Taking into account the prior contribution, the combined functional is given by
\begin{align}
\mathscr{R}_\rho\big(\bm{\rho},\sdist,\alpha,\beta\big) \equiv \mathscr{S} + \frac{1}{2}\sum^{4d}_{j=1}\norm{\rho_j-\bar{\rho}_j}^2_{\mathcal{C}_{\rho_j}(I)}\quad,
\end{align}
where $\mathcal{C}_{\rho_j} = \widetilde{\sigma}_{{\rho_j}}^2 (C * \cdot)$, $\widetilde{\sigma}_{{\rho_j}}=\xi_\rho \sigma_{{\rho_j}}$, and $\xi_\rho \in \R$ is a user-selected parameter that determines the confidence in the prior assumption.\\

\subsubsection{Consistency with measured $k$-space signal}
We consider the noise in the sparse $k$-space signals $s^\star_j$ to be white and Gaussian, with zero mean and standard deviation $\sigma_j$. The discrepancy between the measured sparse signals $s^\star_j$ and a signal $s_j \equiv \mathcal{F}(\rho_je^{i\varphi_j})$ is thus given by
\begin{align}
\mathscr{E}(\bm{\varphi},\bm{\rho}) \equiv \sum^{4d}_{j=1}\frac{1}{2\sigma_j^2}\norm{s^\star_j-\mathcal{P}\mathcal{F}(\rho_je^{i\varphi_j})}^2_{L^2(I_s)}\quad,
\label{eq:kspace_rec_consistency}
\end{align}
with $\rho_je^{i\varphi_j} \equiv w_j$ being the complex image.\\

\begin{figure}
\centering
\includegraphics[width=\linewidth]{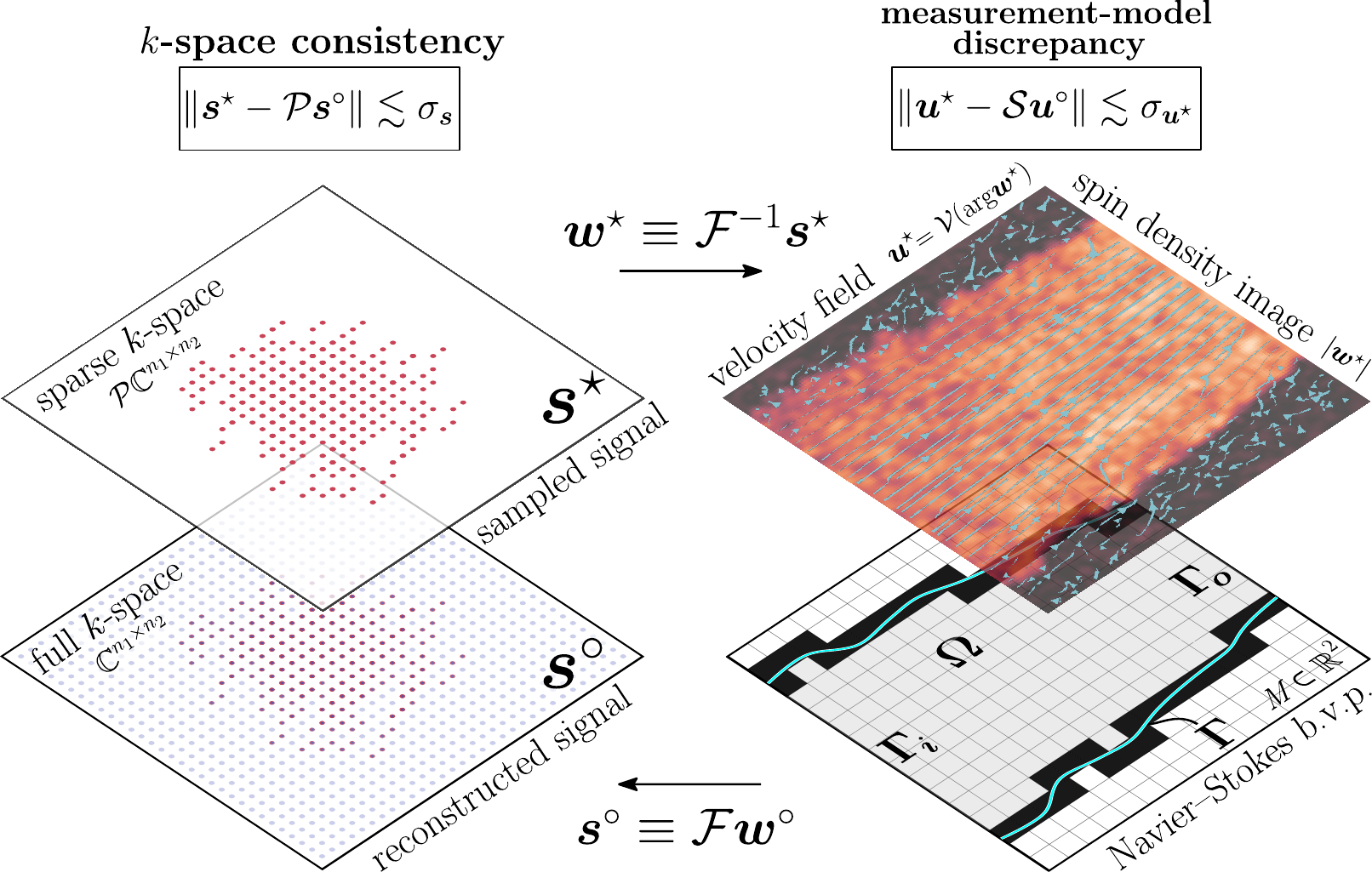}
\label{fig:pics2d}
\caption{A set of sparse $k$-space signals ($\bm{s}^\star$) can be zero-filled and Fourier-transformed to the physical space in order to produce a set of complex images ($\bm{w}^\star$), which provide a first estimate for the measured velocity field ($\bm{u}^\star$) and the ${}^1$H spin density inside an object. A Navier--Stokes problem is then used to reconstruct the measured velocity field and the boundary of the object. The reconstructed complex image ($\bm{w}^\circ$) is then assembled and transformed back to $k$-space ($\bm{s}^\circ$), where it can be compared with the original sparse signal.}
\end{figure}

\subsubsection{The inverse problem}
We now collect the above results to formulate an inverse problem where the reconstructed phases $\bm{\varphi}^\circ$, N--S parameters $\bm{x}^\circ$, magnitudes $\bm{\rho}^\circ$, and segmentation constants $\bm{c}^\circ=(\alpha^\circ,\beta^\circ)$, are the minimizers of the following nonlinearly-constrained optimization problem
\begin{align}
\label{eq:inv_prob_constrained}
\big(\bm{\varphi}^\circ,\bm{x}^\circ,\bm{\rho}^\circ,&\ \bm{c}^\circ\big) = \underset{\bm{\varphi},\bm{x},\bm{\rho},\bm{c}}{\textrm{argmin}}\ \big(\mathscr{R}_{\varphi} + \mathscr{R}_{x} + \mathscr{R}_{\rho}\big) \\ 
\text{subject to}\quad& \text{i)}\quad \rev{\norm{s^\star_j-\mathcal{P}\mathcal{F}\big(\rho_je^{i\varphi_j}\big)}_{L^2(I_s)} \leq \sqrt{\abs{I_s}} \varepsilon_j\ \text{for all}\ j \nonumber}\\
\quad& \text{ii)}\quad \bm{u} = \mathcal{Z}\ \bm{x} \nonumber\quad,
\end{align}
where $\varepsilon_j \propto \sigma_j$ is a user-selected value proportional to the standard deviation of noise. We observe that the N--S problem \eqref{eq:navierstokes_bvp}, encoded in the operator $\mathcal{Z}$, acts as a nonlinear equality constraint in order to ensure that $\bm{u}$ is always a N--S solution, which is uniquely defined by the parameters $\bm{x}$ ($\bm{u}^\circ = \mathcal{Z}\bm{x}^\circ$).\\

\subsubsection{Relaxing the constraints}
We recast problem \eqref{eq:inv_prob_constrained} into a variational form by defining the Lagrangian
\begin{align}
\mathscr{J} \equiv \mathscr{R}_{\varphi} + \mathscr{R}_{x} + \mathscr{R}_{\rho} + \mathscr{E} + \mathscr{M}\quad,
\label{eq:lagrangian}
\end{align}
where $\mathscr{M}$ is the weak formulation of the N--S problem \eqref{eq:navierstokes_bvp}, for test functions $(\bm{v},q) \in \bm{H}^1(\Omega)\times L^2(\Omega)$, given by

\begin{align}
\mathscr{M}(\bm{u},p&,\bm{v},q;\bm{x}) \equiv \int_\Omega\Big(\bm{v}\bm{\cdot}\big( \bm{u}\bm{\cdot}\nabla\bm{u}\big) + \nu\nabla\bm{v}\bm{:}\nabla \bm{u} -(\nabla\bm{\cdot}\bm{v}) p \nonumber \\ & - q(\nabla\bm{\cdot} \bm{u})\Big) +\int_{\Gamma_o}\bm{v}\bm{\cdot}\bm{g}_o +\int_{\Gamma\cup\Gamma_i} \bm{v}\bm{\cdot}(-\nu\partial_{\bm{\nu}}\bm{u}+p\bm{\nu}) \nonumber\\ &+ \mathscr{N}_{\Gamma_i}(\bm{v},q,\bm{u};\bm{g}_i)+\mathscr{N}_{\Gamma}(\bm{v},q,\bm{u};\bm{0})\quad,
\label{eq:navier_stokes_weak}
\end{align}
and $\mathscr{N}$ is the Nitsche penalty term \cite{Nitsche1971} 
\begin{gather}
\mathscr{N}_{\Gamma}(\bm{v},q,\bm{u};\bm{g}) \equiv \int_{\Gamma} (-\nu\partial_{\bm{\nu}}\bm{v}+q\bm{\nu}+ \eta\bm{v})\bm{\cdot}(\bm{u}-\bm{g})\quad,
\end{gather}
which we use to weakly impose a Dirichlet boundary condition $\bm{g} \in \bm{L}^2(\Gamma)$ on a boundary $\Gamma$, for fixed penalty $\eta \in \mathbb{R}$ \cite[Section~2.7]{Kontogiannis2021}.\\

\subsubsection{Euler--Lagrange equations}
Instead of solving \eqref{eq:inv_prob_constrained} as is, it is preferable to search for critical points of \eqref{eq:lagrangian} by solving the nonlinear Euler--Lagrange (E--L) system for the unknowns
\begin{equation*}
\bm{v},\ q,\ \bm{u},\ p,\ \bm{\varphi},\ \bm{x},\ \bm{\rho},\ \bm{c}\ ,
\end{equation*}
where $\bm{v}$ is the adjoint velocity and $q$ is the adjoint pressure, acting as Lagrange multipliers in \eqref{eq:lagrangian} via $\mathscr{M}$. We find that the E--L equations
\begin{equation}
\delta_{\bm{v}}\mathscr{J} = \delta_{\bm{v}}\mathscr{M} = \bm{0} \quad,\quad \delta_{q}\mathscr{J} = \delta_{q}\mathscr{M} = 0\quad,
\end{equation}
are in fact the N--S equations and the weakly-enforced boundary conditions, i.e. problem \eqref{eq:navierstokes_bvp}. The next set of equations, namely
\begin{equation}
\delta_{\bm{u}}\mathscr{J} = \delta_{\bm{u}}\big(\mathscr{U}+\mathscr{M}\big) = \bm{0} \quad,\quad \delta_{p}\mathscr{J} = \delta_{p}\mathscr{M} = 0\quad,
\end{equation}
gives rise to the adjoint N--S problem \cite[Section~2.3.1]{Kontogiannis2021}
\begin{gather}
\left\{\begin{alignedat}{2}
- \bm{u}\bm{\cdot}\big(\nabla\bm{v} + (\nabla\bm{v})^\dagger\big) -\nu\Delta \bm{v} + \nabla q &= -D_{\bm{u}}\mathscr{J} \quad &&\textrm{in}\quad \Omega \\ 
\nabla \bm{\cdot} \bm{v} &= 0 \quad &&\textrm{in}\quad \Omega\\
\bm{v} &= \bm{0} \quad &&\textrm{on}\quad \Gamma\cup\Gamma_i\\
{(\bm{u}\bm{\cdot}\bm{\nu})\bm{v}}+{(\bm{u}\bm{\cdot}\bm{v})\bm{\nu}}+\nu\partial_{\bm{\nu}}\bm{v}-q\bm{\nu} &= \bm{0} \quad &&\textrm{on}\quad \Gamma_o
\end{alignedat}\right.
\label{eq:navier-stokes_adjoint_problem}
\end{gather}
Note that
\begin{equation}
D_{\bm{u}} \mathscr{J} = D_{\bm{u}} \mathscr{U} =  -\mathcal{S}^\dagger \mathcal{C}_{\bm{u}}^{-1}\big(\bm{u}^\star-\mathcal{S}\bm{u}\big)\quad,
\end{equation}
and that both $\bm{v}$ and $q$ vanish when $D_{\bm{u}} \mathscr{U} \equiv \bm{0}$ in $I_w$, i.e. when the measured velocity is identical to the modelled velocity in the user-selected area of interest $I_w$. The rest of the equations comprising the E--L system are
\begin{alignat}{2}
\delta_{\bm{\varphi}}\mathscr{J} &= \delta_{\bm{\varphi}}\big(\mathscr{R}_\varphi + \mathscr{E}\big) &&= \bm{0} \label{eq:el_phase}\\
\delta_{\bm{x}}\mathscr{J} &= \delta_{\bm{x}}\big(\mathscr{R}_x+\mathscr{S}+\mathscr{M}\big) &&= \bm{0} \label{eq:el_ns_unk}\\
\delta_{\bm{\rho}}\mathscr{J} &= \delta_{\bm{\rho}}\big(\mathscr{R}_\rho + \mathscr{E}\big) &&= \bm{0} \label{eq:el_mag}\\
\delta_{\bm{c}}\mathscr{J} &= \delta_{\bm{c}}\mathscr{S} &&= \bm{0}\quad \label{eq:el_segm},
\end{alignat}
which we expand in Appendix \ref{app:sadirections} to obtain an explicit formula for the steepest ascent direction of each unknown. We then use these directions in order to find a critical point of $\mathscr{J}$ using an optimization algorithm.\\

\subsubsection{Solving the E--L system}
For large scale problems such as \eqref{eq:inv_prob_constrained} it is prohibitively expensive to solve the full, implicit \mbox{E--L} system. Instead, we devise a segregated method that solves this {\color{black} nonlinear and nonconvex} system, which is briefly explained in \mbox{algorithm \ref{algo:pics}}. Each iteration in the main loop of algorithm \ref{algo:pics} consists of four main stages:
\begin{enumerate}
 \item First, since $\mathscr{J} = \mathscr{J}(\bm{\varphi},\bm{x},\bm{\rho},\bm{c})$, we fix the phases $\bm{\varphi}$, the magnitudes $\bm{\rho}$, and the segmentation constants $\bm{c}$, and update the unknown N--S parameters $\bm{x}$ by solving one iteration of the inverse N--S problem. This stage is actually identical to the algorithm presented in \cite{Kontogiannis2021}, which was devised for the reconstruction of noisy velocity images (fully-sampled PC-MRI signals), where $\bm{\varphi},\bm{\rho}$ are always constant, and the segmentation functional $\mathscr{S}$ is absent, i.e. $\bm{c}$ is not involved. The updated N--S parameters $\bm{x}_{k+1}$ are then used to update the modelled velocity field to $\bm{u}_{k+1} = \mathcal{Z}^{\bm{u}_k}\bm{x}_{k+1}$, where $\mathcal{Z}^{\bm{u}_k}$ denotes the Oseen linearization of the N--S problem \eqref{eq:navierstokes_bvp} around $\bm{u}_k$. The Oseen linearization around $\bm{u}_k$ replaces the nonlinear convective term $\bm{u}\bm{\cdot}\nabla\bm{u}$ in \eqref{eq:navierstokes_bvp} with the linear convective term $\bm{u}_k\bm{\cdot}\nabla\bm{u}$. 
 \item Next, we fix $\bm{x},\bm{\rho}$ and $\bm{c}$ and reconstruct the phases, drawing information from the modelled velocity field $\bm{u}_{k+1}$, \textit{a priori} phase information, and the $k$-space signals. 
 \item We then fix $\bm{\varphi},\bm{x}$ and $\bm{\rho}$ in order to compute the segmentation constants $\alpha,\beta$, given by the explicit formulas \eqref{eq:alpha_value} and \eqref{eq:beta_value}.
 \item Finally, we fix $\bm{\varphi},\bm{x}$ and $\bm{c}$ in order to reconstruct the magnitudes, drawing information from the energy-based segmentation functional, \textit{a priori} magnitude information, and the $k$-space signals.
\end{enumerate}
During the first stage of each iteration, we further update an inverse Hessian approximation of the unknown parameters $\bm{x}$ using the damped BFGS quasi-Newton method \cite{Fletcher2000,Nocedal2006}, and this allows us to estimate the uncertainty in the predicted shape, the boundary conditions, and the kinematic viscosity \cite[Section~2.6]{Kontogiannis2021}. {\color{black} Algorithm \ref{algo:pics} terminates if either the reconstructed velocity, phases, and magnitudes are consistent with the data and the update for the unknowns $\bm{\varphi},\bm{x},\bm{\rho},\bm{c}$ is below the user-specified tolerance, or the line search of every individual stage fails to reduce $\mathscr{J}$ further. The reconstructed velocity components are consistent with the data when the discrepancy between the velocity component and the respective phase difference, $\mathscr{U}_j$, i.e. the summands of equation \eqref{eq:velocity_phase_diff_discrepancy}, satisfy $\sqrt{2\mathscr{U}_j/\abs{I_s}} < \varepsilon_j$ for every $j$, where $\varepsilon_j \lesssim 1$. The reconstructed phases and magnitudes are consistent with the data when the \mbox{$k$-space} consistency norms of the scans, $\mathscr{E}_j$, i.e. the summands of equation \eqref{eq:kspace_rec_consistency}, satisfy $\sqrt{2\mathscr{E}_j/\abs{I_s}} < \varepsilon_j$ for every $j$. Every line search starts with a step size of $\tau = 1$, and the step size is halved until $\mathscr{J}_{k+1} < \mathscr{J}_k$. It is often the case that the search of the N--S parameters (first stage of algorithm \ref{algo:pics}) converges faster than the reconstruction of the phases and the magnitudes. In this case, the inverse N--S problem within the loop does not need to be solved further, and the reconstructed modelled velocity field (given by the converged N--S parameters) is used to reconstruct the phases and the magnitudes until all stages of the algorithm have converged. 

The reconstruction problem \eqref{eq:inv_prob_constrained} that algorithm \ref{algo:pics} solves is nonlinear and nonconvex because it involves a moving domain in which a N--S problem is solved, a $k$-space signal that is decomposed into phase and magnitude components, and a $2\pi$-periodic prior for the phases. Ill-posedness is largely mitigated using a Bayesian regularization framework. The well-posedness of Bayesian inverse N--S problems is addressed in \cite{Cotter2009}. Problem \eqref{eq:inv_prob_constrained}, however, is more complicated than an inverse N--S problem alone. Nevertheless, for the test case that we have studied here, we observe that algorithm \ref{algo:pics} successfully recovers the true solution, as it is shown in table \ref{tab:sampling_density_effect}.}

\begin{algorithm}[h]
\caption{PICS for sparse PC-MRI signals.} \label{algo:pics}
\textbf{Input:} sparse PC-MRI signals $\bm{s}^\star$, noise variances, initial guesses (priors) for the unknowns $\bar{\bm{\varphi}},\bar{\bm{x}},\bar{\bm{\rho}}$\\
\Begin{
$k \leftarrow 0$\\
\textbf{Initialization}\\
\makebox[0pt][l]{$\bm{u}^\star_0$}\phantom{banana} $\leftarrow$ {zero-filled velocity from $\bar{\bm{\varphi}}$ - eq. \eqref{eq:phase_diff}}\\
\makebox[0pt][l]{$\bar{\phi}_\pm$}\phantom{banana} $\leftarrow$ {signed distance field from $\sum_j\bar{\rho}_j$ \cite{Kontogiannis2021}}\\
\makebox[0pt][l]{$(\bm{u},p)_0$}\phantom{banana} $\leftarrow$ {$\mathcal{Z}\ \bar{\bm{x}}$ - N--S problem \eqref{eq:navierstokes_bvp}}\\[4pt]
\While{\footnotesize \tt{convergence\_criterion\_is\_not\_met}\\[1.5pt]}
{
\textbf{1) inverse N--S - $\textrm{min}$}\ $\mathscr{J}({\color{gray}\bm{\varphi}_k},\bm{x},{\color{gray}\bm{\rho}_k},{\color{gray}\bm{c}_{k}})$:\\
\makebox[0pt][l]{$(\bm{v},q)_k$}\phantom{avocados} $\leftarrow$ adjoint N--S problem \eqref{eq:navier-stokes_adjoint_problem}\\
\makebox[0pt][l]{$\widehat{D}_{\bm{x}}\mathscr{J}$}\phantom{avocados} $\leftarrow $ s. a. directions \eqref{eq:gi_grad}-\eqref{eq:geomflow_segm}\\
\makebox[0pt][l]{$\widetilde{H}^{k}_x,\tau$}\phantom{avocados} $\leftarrow$ {approx. inv. Hessian and l. s. \cite{Kontogiannis2021}}\\
\makebox[0pt][l]{$\bm{x}_{k+1}$}\phantom{avocados} $\leftarrow$ $\bm{x}_k -\tau\ \widetilde{H}^{k}_x\widehat{D}_{\bm{x}}\mathscr{J}$ \\
\makebox[0pt][l]{$(\bm{u},p)_{k+1}$}\phantom{avocados} $\leftarrow$ $\mathcal{Z}^{\bm{u}_k}\bm{x}_{k+1}$ - lin. N--S problem \eqref{eq:navierstokes_bvp}\\[3pt]
\vspace{0.1cm}\hrule\vspace{0.1cm}
\textbf{2) phase - $\textrm{min}$}\ $\mathscr{J}(\bm{\varphi},{\color{gray}\bm{x}_{k+1}},{\color{gray}\bm{\rho}_k},{\color{gray}\bm{c}_{k}})$:\\
\makebox[0pt][l]{$\widehat{D}_{\bm{\varphi}}\mathscr{J},\tau$}\phantom{avocados} $\leftarrow $ s. a. directions \eqref{eq:phase_grad} and l. s.\\
\makebox[0pt][l]{$\bm{\varphi}_{k+1}$}\phantom{avocados} $\leftarrow$ {$\bm{\varphi}_k-\tau\ \widehat{D}_{\bm{\varphi}}\mathscr{J}$}\\
\makebox[0pt][l]{$\bm{u}^\star_{k+1}$}\phantom{avocados} $\leftarrow$ {measured vel. from $\bm{\varphi}_{k+1}$ - eq. \eqref{eq:phase_diff}}\\[3pt]
\vspace{0.1cm}\hrule\vspace{0.1cm}
\textbf{3) segmentation - $\textrm{min}$}\ $\mathscr{J}({\color{gray}\bm{\varphi}_{k+1}},{\color{gray}\bm{x}_{k+1}},{\color{gray}\bm{\rho}_k},\bm{c})$:\\
{\color{black}\makebox[0pt][l]{$\bm{c}_{k+1}$}\phantom{avocados} $\leftarrow$ {compute $(\alpha,\beta)_{k+1}$ - eq. \eqref{eq:alpha_value},\eqref{eq:beta_value}}}\\[3pt]
\vspace{0.1cm}\hrule\vspace{0.1cm}
\textbf{4) magnitude - $\textrm{min}$}\ $\mathscr{J}({\color{gray}\bm{\varphi}_{k+1}},{\color{gray}\bm{x}_{k+1}},\bm{\rho},{\color{gray}\bm{c}_{k+1}})$:\\
\makebox[0pt][l]{$\widehat{D}_{\bm{\rho}}\mathscr{J},\tau$}\phantom{avocados} $\leftarrow $ s. a. directions \eqref{eq:mag_grad} and l. s.\\
\makebox[0pt][l]{$\bm{\rho}_{k+1}$}\phantom{avocados} $\leftarrow$ {$\bm{\rho}_k-\tau\ \widehat{D}_{\bm{\rho}}\mathscr{J}$}\\[3pt]
\vspace{0.1cm}\hrule\vspace{0.1cm}
$k \leftarrow k+1$
}
\textbf{Output:}\\
\makebox[0pt][l]{$\bm{x}^\circ$}\phantom{avocados}$\leftarrow$ \makebox[0pt][l]{$\bm{x}_k$}\phantom{bananas} (inferred N--S parameters)\\
\makebox[0pt][l]{$(\bm{u}^\circ,p^\circ)$}\phantom{avocados}$\leftarrow$ \makebox[0pt][l]{$(\bm{u},p)_k$}\phantom{bananas} (N--S velocity and pressure)\\
\makebox[0pt][l]{$(\bm{\varphi}^\circ,\bm{\rho}^\circ)$}\phantom{avocados}$\leftarrow$ \makebox[0pt][l]{$(\bm{\varphi},\bm{\rho})_k$}\phantom{bananas} (reconstructed phases/mag.)\\[3pt]
\textbf{Optional output:}\\
$\gamma_w \leftarrow$ wall shear rate from $\bm{u}^\circ$, $\bm{x}^\circ$
}
\text{\footnotesize \color{gray} s. a.: steepest ascent, l. s.: line search, lin.: linearized}
\end{algorithm}

\subsubsection{Numerics}
To solve boundary value problems, problem \eqref{eq:navierstokes_bvp} and \eqref{eq:navier-stokes_adjoint_problem}, for example, we use an immersed boundary finite element method, the details of which can be found in \cite[Section~2.7]{Kontogiannis2021}. We implement algorithm \ref{algo:pics}, and all the necessary numerical methods, in Python modules, using Python's standard libraries for scientific computing: NumPy \cite{harris2020array} and SciPy \cite{2020SciPy-NMeth}.

\section{Sparse PC-MRI reconstruction using PICS}
\label{sec:rec_results_main}
\begin{figure}
\centering
\includegraphics[width=0.49\textwidth]{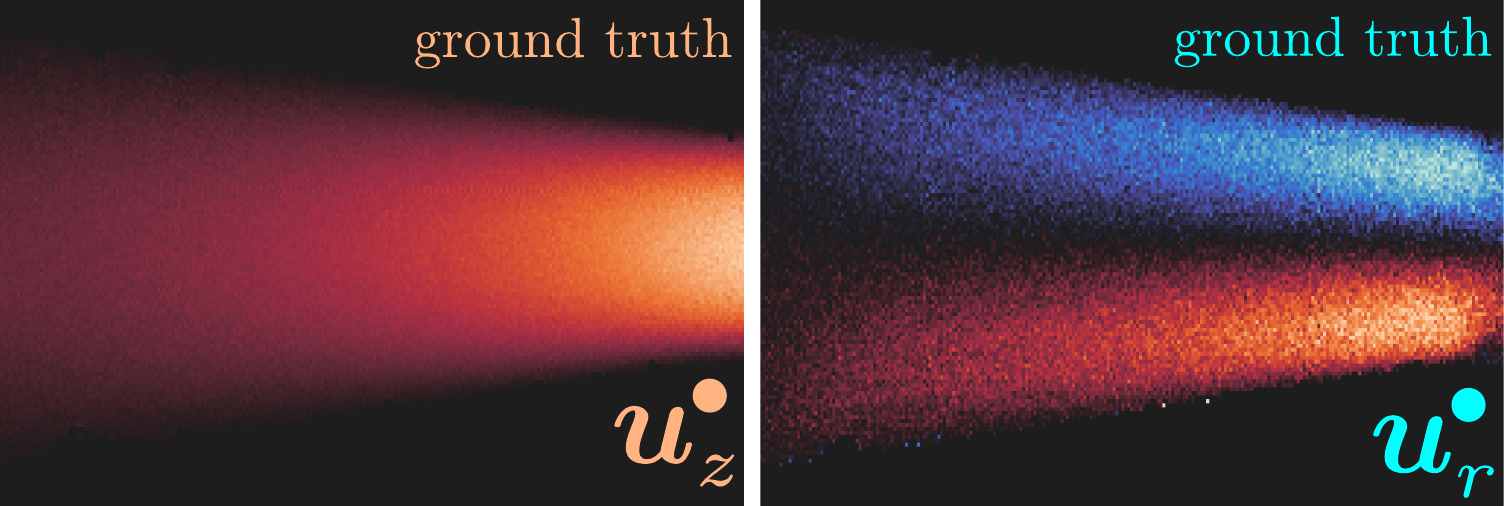}
\caption{Ground truth for the axial velocity component $u^\bullet_z$ ($\text{SNR}_z\simeq52$), and the radial velocity component $u^\bullet_r$ ($\text{SNR}_r\simeq43$), for the left-to-right flow through the converging nozzle (colormap scale shown in figure \ref{fig:rec_results_1}). The images were obtained using phase-contrast MRI (see Appendix \ref{app:pcmri_exp}).}
\label{fig:gt_images}
\end{figure}

We now apply algorithm \ref{algo:pics} to sparse and noisy (low SNR) PC-MRI signals of an axisymmetric flow of water/glycerol through a converging nozzle \cite{Kontogiannis2021}. The description of the \mbox{PC-MRI} experiment can be found in Appendix \ref{app:pcmri_exp}. \rev{Since the $k$-space was fully-sampled, we sparsify the $k$-space signals using two different sparse sampling patterns $\mathcal{P}$: i) a sampling pattern $\mathcal{P}_\odot$, created from a two-dimensional normal distribution, and ii) a sampling pattern $\mathcal{P}_\parallel$, created from a one-dimensional normal distribution of lines. The sampling pattern $\mathcal{P}_\parallel$ is commonly used in MRI because conventional acquisition protocols sample $k$-space lines for each system excitation.} For these sparse sampling patterns, we further investigate the effect of subsampling on the velocity reconstruction error. The reconstruction error of a modelled velocity component $u^\circ_i$ is measured by
\begin{equation}
\mathcal{E}^\bullet_{u^\circ_i} = (\sqrt{\abs{M}}\sigma_{u^\bullet_i})^{-1} \norm{u^\bullet_i-u^\circ_i}_{L^2(M)}\quad, 
\label{eq:rec_error}  
\end{equation}
where $u^\bullet_i$ is the corresponding ground truth velocity image, which, in this paper, is a high signal-to-noise ratio image of the same flow (figure \ref{fig:gt_images}), and $\sigma_{u^\bullet_i}^2$ is the estimated variance of Gaussian white noise in the ground truth image.

\subsection{Noise in the signal, the phase, and the magnitude}
The noise in each $k$-space signal $s^\star_j$ is assumed to be Gaussian, white, and additive, with zero mean and variance $\sigma^2_j$. It is common practice in MRI to estimate the noise from regions of the magnitude images with no signal. For fully-sampled signals, the magnitude of the signal $s^\star_j$, given by $\rho_j\equiv\abs{\mathcal{F}^{-1}s^\star_j}$, is distributed according to a Rayleigh distribution \cite{Gudbjartsson1995}. The noise variance in $s^\star_j$ is thus given by
\begin{equation}
\sigma^2_j = 2(4-\pi)^{-1}\ \sigma^2_{\rho_j}\quad,
\end{equation}
where $\sigma^2_{\rho_j}$ is the noise variance of the magnitude image $\rho_j$, estimated from a region with no signal. The SNR of the $j$-th signal is then computed from
\begin{equation}
\text{SNR}_j = \frac{\mu_{\rho_j}}{\sigma_{\rho_j}}\quad,
\end{equation}
where $\mu_{\rho_j}$ is the mean of $\rho_j$, taking into account only the active regions (non-zero signal). When $\text{SNR}_j > 3$, the noise in the phase images $\varphi_j$ approximates a zero-mean Gaussian distribution with variance \cite{Gudbjartsson1995}
\begin{equation}
\sigma^2_{\varphi_j} = \text{SNR}_j^{-2} \quad.
\end{equation}
Considering that the measured velocity, given by equation \eqref{eq:phase_diff}, is computed from the phase differences, the noise variance of the $k$-th velocity component is estimated by
\begin{equation}
\sigma^2_{u_k} = c^2_k \sum_{j} \sigma^2_{\varphi_j} \quad,
\label{eq:vel_noise_fs}
\end{equation}
for $j=1,\dots,4d$ and $k\equiv(j-1)\textrm{div}4 +1$, where $\textrm{div}$ denotes integer division.

\begin{figure}
\centering
\includegraphics[width=0.49\textwidth]{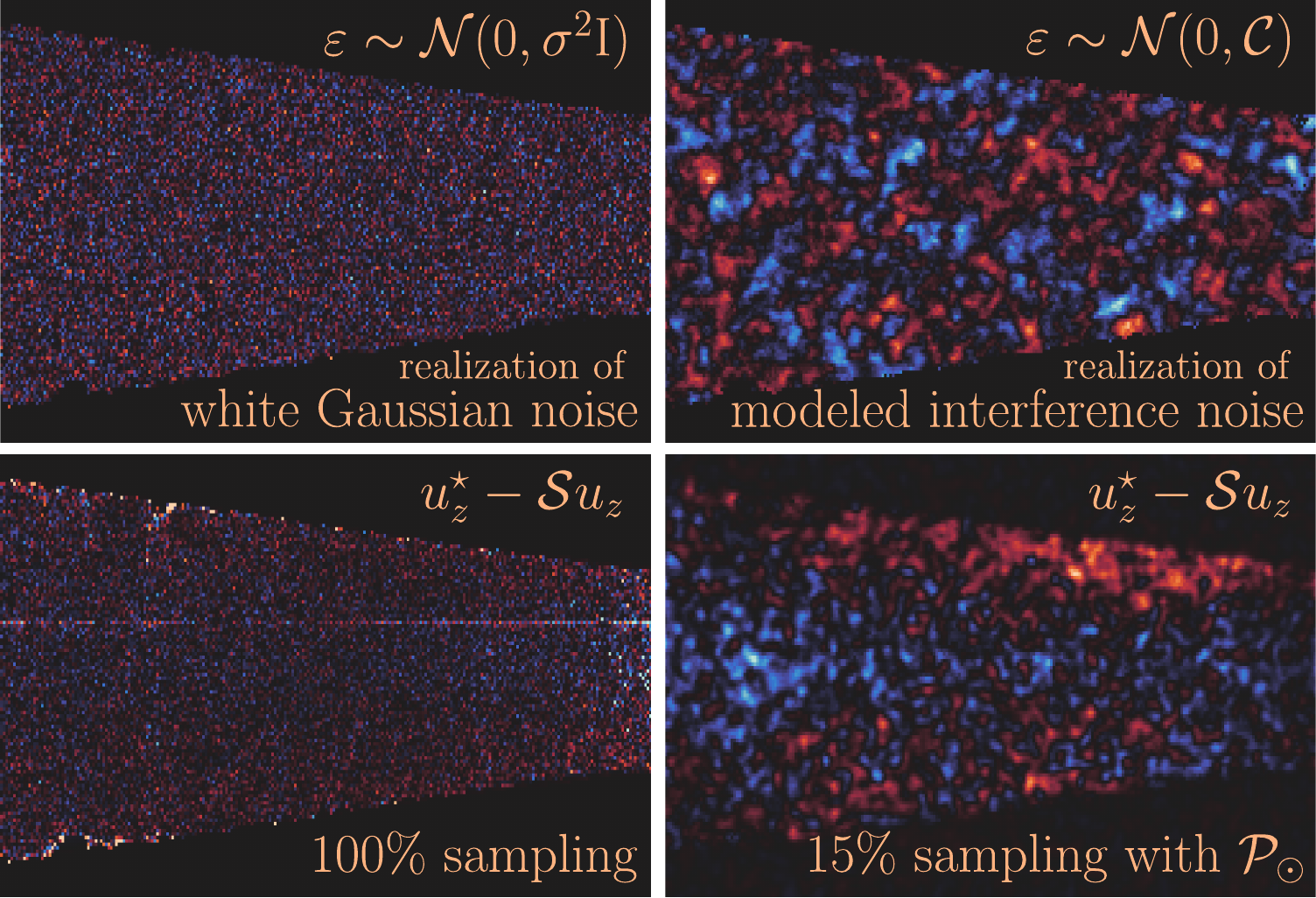}
\caption{Left frames: velocity discrepancy for fully-sampled (100\% \mbox{$k$-space} coverage) signals. The noise $\varepsilon$ in the measured velocity can be considered to be white and Gaussian when the signals are fully-sampled, i.e. \mbox{$u^\star_k - \mathcal{S}u_k \sim \mathcal{N}(0,\sigma^2\mathrm{I})$}. Right frames: velocity discrepancy for sparsely-sampled signals (15\% $k$-space coverage with $\mathcal{P}_\odot$). For sparsely-sampled signals we observe correlations that depend on the sampling pattern $\mathcal{P}$. To model these correlations, we introduce a covariance operator $\mathcal{C}$ that is constructed using an exponential kernel, i.e. \mbox{$u^\star_k - \mathcal{S}u_k \sim \mathcal{N}(0,\mathcal{C})$}.} 
\label{fig:noise_modeling}
\end{figure}

Based on the above, for fully-sampled $k$-space signals the noise $\varepsilon_{u_k}$ in the velocity image of the $k$-th component is distributed according to $\mathcal{N}(0,\sigma_{u_k}^2\mathrm{I})$, \rev{where $\mathcal{N}(m,\mathcal{C})$ denotes the normal distribution with mean $m$ and covariance operator $\mathcal{C}$}. For sparsely-sampled \mbox{$k$-space} signals we observe that $\varepsilon_{u_k}$ is correlated. Here, we model this correlation using an exponential covariance function (see section \ref{subsec:phase_reg}). Figure \ref{fig:noise_modeling} shows the velocity discrepancy ($u^\star_z - \mathcal{S}u_z$) between the measured ($u^\star_z$) and the modelled ($u_z$) axial velocity components for fully-sampled signals, and 15\% sparsely-sampled signals using $\mathcal{P}_\odot$ \rev{(see section \ref{sec:sampling_patterns})}. For fully-sampled \mbox{$k$-spaces}, we observe that
\begin{equation}
u^\star_k - \mathcal{S}u_k \sim \mathcal{N}(0,\sigma^2_{u_k}\mathrm{I})\quad,
\end{equation}
and this justifies the use of the norm $\sigma^{-2}_{u_k}\norm{u^\star_k - \mathcal{S}u_k}^2_{L^2}$ in \cite{Kontogiannis2021}, where fully-sampled signals were considered.
For sparsely-sampled $k$-spaces, we observe that the noise is correlated and we assume that
\begin{equation}
u^\star_k - \mathcal{S}u_k \sim \mathcal{N}(0,\mathcal{C}_{u_k})\quad,
\end{equation}
where $\mathcal{C}_{u_k}$ is given by equation \eqref{eq:exp_cov_op}, with characteristic length $\ell$ equal to the smallest resolved length scale in the image $I$.

\begin{table}
\centering
  \caption{Noise level for the acquired noisy signals and the resulting velocity images.}
  \label{tab:noise_sigmas}
    \begin{tabular}{cccccccc}
        \multicolumn{8}{c}{$k$-space signals $s^\star_j$}\\\hline
        $\sigma_1$ & $\sigma_2$ & $\sigma_3$ & $\sigma_4$ & $\sigma_5$ & $\sigma_6$ & $\sigma_7$ & $\sigma_8$\\
        38.4 & 38.5 & 14.7 & 14.6 & 43.1 & 42.9 & 19.2 & 19.2 
  \end{tabular}
  \begin{tabular}{lccr}
        \\[1.5pt]
        vel. component & $c_k$ [cm/s] & $\sigma_{u_{(\cdot)}}$ [cm/s] & SNR$_j$ \\
        \hline\\ 
        axial $z$ ($j=1,\dots4$)  & 1.988 & 0.348 & 11.4 \\[3pt]
        radial $r$ ($j=5,\dots8$) & 0.529 & 0.113 &  9.3
  \end{tabular}
\end{table}

\subsection{Generating the sparse sampling patterns $\mathcal{P}_\odot$ and $\mathcal{P}_\parallel$}
\label{sec:sampling_patterns} 
To generate the sampling pattern $\mathcal{P}_\odot$ for a discretized two-dimensional \mbox{$k$-space}, we draw random samples $\bm{y}^s_k$ from a normal distribution $\mathcal{N}(\bm{\mu},\bm{\Sigma})$, whose mean and covariance are given by
\begin{equation}
\bm{\mu} = \Big(\frac{n_1}{2},\frac{n_2}{2}\Big)\quad,\quad \bm{\Sigma} = \textrm{diag}\Big(\frac{\omega_1n_1}{4},\frac{\omega_2n_2}{4}\Big)\quad,
\end{equation} 
where $n_i$ denotes the number of points (or pixels), and $\omega_i$ denotes the coverage, in the sense of a $2\sigma$ interval, along the $i$-th direction. We further round the samples \mbox{$\bm{y}^s_k = (y^s_{k1},y^s_{k2}) \mapsto \nint{\bm{y}^s_k} = (\nint{y^s_{k1}},\nint{y^s_{k2}})$} to the closest integer, so that they correspond to the pixel index. The degree of subsampling is controlled by the sampling density, which is defined by $N_s/N$, where $N_s$ is the number of admissible sampled \mbox{$k$-space} points (we reject samples that either repeat on the same pixel or lie outside the $k$-space domain) and $N=n_1n_2$ is the total number of $k$-space points. The sampling pattern $\mathcal{P}_\odot: \mathbb{Z}^{n_1\times n_2} \to \R$ is then given by
\begin{equation}
\mathcal{P}_\odot(\bm{y}) = 1 \quad \text{if}\quad \bm{y} = \nint{\bm{y}^s_k} \quad \text{else}\quad 0 \ ,
\end{equation}
for $k = 1,\dots,N_s$. The sampling pattern $\mathcal{P}_\parallel: \mathbb{Z}^{n_1\times n_2} \to \R$ consists of random samples of vertical lines, and is given by
\begin{equation}
\mathcal{P}_\parallel(y_1,\cdot) = 1 \quad \text{if}\quad y_1 = \nint{{y}^s_{k1}} \quad \text{else}\quad 0 \ ,
\end{equation}
for $k = 1,\dots,N_s$.

\subsection{Reconstructing sparse PC-MRI signals}
\label{sec:reconstruction}
\rev{Using the sparse sampling patterns $\mathcal{P}_\odot$ and $\mathcal{P}_\parallel$, with \mbox{$\omega_1 = \omega_2 = 0.35$}, and for various sampling densities \mbox{$N_s/N$}, we sparsify the originally full $k$-space signals that we have acquired for the flow through the nozzle.} We use \mbox{algorithm \ref{algo:pics}} to reconstruct the sparse, noisy (low SNR) signals $\bm{s}^\star$ (see table \ref{tab:noise_sigmas} for the noise level), and compare the results with the ground truth (high SNR) images that we have acquired for the same flow (figure \ref{fig:gt_images}). The low SNR images required a total scanning time of 2.6 minutes per velocity image (axial and radial components), and the high SNR images required a total scanning time of 68 minutes per velocity image. These scanning times correspond to 100\% $k$-space coverage. Thus, all other things being equal, the 15\%-sampled low SNR signals would have been acquired in $\sim23$ seconds, assuming that acquisition time linearly scales with sampling density\footnote{The validity of this assumption depends on both the sparse sampling pattern and the pulse sequence of the magnetic resonance experiment.}.

\subsubsection{Input and initialization}
The input set of \mbox{algorithm \ref{algo:pics}} consists of the $k$-space signals $\bm{s}^\star$, the estimated noise variances (table \ref{tab:noise_sigmas}), and the prior assumption for $\bm{\varphi},\bm{\rho}$ and $\bm{x}$. The prior assumptions are in fact Gaussian random fields, and we therefore need to define both their mean and their variance (confidence level of \textit{a priori} knowledge). We choose the prior mean of the phase ($\bar{\bm{\varphi}}$) and the magnitude ($\bar{\bm{\rho}}$) images to be their respective zero-filling solution. \rev{For the N--S unknowns, we choose the prior mean of the signed distance function (SDF) ($\bar{\phi}_\pm$) to be the SDF that corresponds to the Chan--Vese segmentation \cite{Chan2001,Getreuer2012} of the averaged zero-filling magnitude image. Shape regularization is particularly important at this point because, for very sparse signals with strong artefacts (e.g. 10\% $\mathcal{P}_\parallel$-sampling), the Chan--Vese magnitude segmentation will provide a poor approximation of the true geometry (we show how we regularize the shape and compute the SDF from a segmentation in \cite[Section~2.4]{Kontogiannis2021})}. We set the prior mean for the inlet velocity boundary condition to $\bar{\bm{g}}_i = (\bar{g}_{iz},0)$, where $\bar{g}_{iz}$ is a parabolic velocity profile that satisfies the zero-velocity boundary condition on the boundary $\partial\Gamma_i$ of the inlet $\Gamma_i$, and has peak velocity equal to $2.5$ cm/s. The prior mean of the outlet boundary condition is $\bar{\bm{g}}_o \equiv \bm{0}$ (pseudotraction-free boundary condition), and of the kinematic viscosity is $\bar{\nu} \simeq 2.54\times10^{-5}$ m$^2$/s \cite{Cheng2008,Volk2018}, which corresponds to a 70\% glycerol in water mixture (see Appendix \ref{app:pcmri_exp}). We explain how we choose the prior variances and the regularization parameters in Appendix \ref{app:priors} \revall{and in \cite[Section~3.6]{Kontogiannis2021}}. Lastly, the initial guess for the measured velocity is obtained using equation \eqref{eq:phase_diff} with $\bar{\bm{\varphi}}$, and the initial guess for the modelled velocity by solving the Navier--Stokes problem $\mathcal{Z}\ \bar{\bm{x}}$.

\begin{table}[!h]
  \caption{Reconstruction error $\mathcal{E}_z^\bullet$\ /\ $\mathcal{E}_r^\bullet$ for each sampling pattern $\mathcal{P}$.}
  \label{tab:sampling_density_effect}
  \begin{tabular}{ccccc}
        & \multicolumn{4}{c}{$k$-space coverage}\\[3pt]
        sampling & 5\% & 10\%  & 15\% & 25\% \\
        \hline \\
        $\mathcal{P}_\odot$ & 1.13\ /\ 0.47 & 0.66\ /\ 0.42  & 0.55\ /\ 0.30 & $\cdot$ \\[3pt]
        $\mathcal{P}_\parallel$ & $\cdot$ & 2.41\ /\ 1.78 & 0.63\ /\ 0.39  & 0.61\ /\ 0.29\\[3pt]
        \hline \\
        full sampling & \multicolumn{4}{c}{0.56\ /\ 0.31}
  \end{tabular}
\end{table}

\subsubsection{Reconstruction results}
\label{sec:rec_results}
\begin{figure*}
\centering
\includegraphics[width=0.9\textwidth]{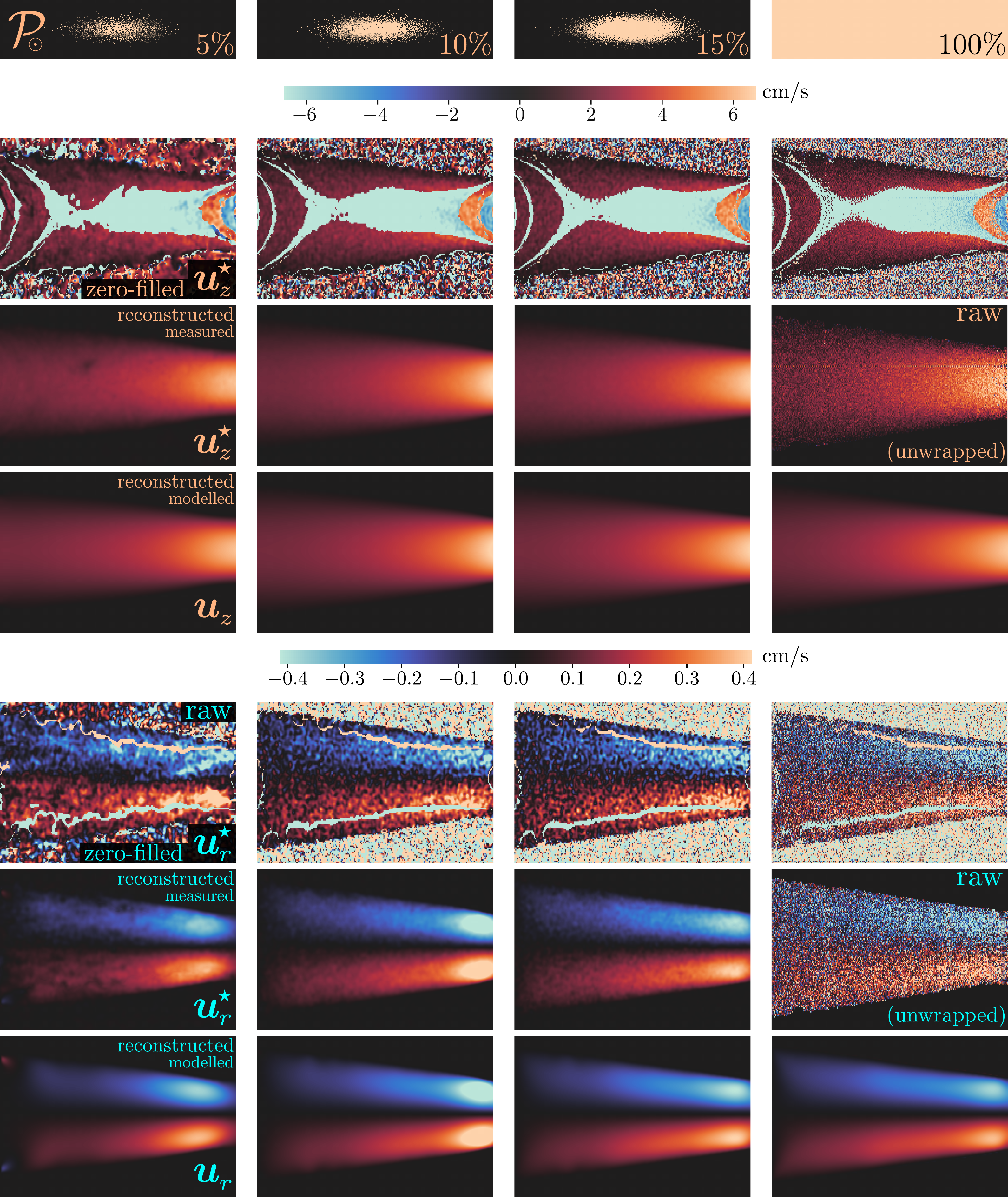}
\caption{We reconstruct sparse $k$-space signals with 5\%, 10\% and 15\% $\mathcal{P}_\odot$-sampling using algorithm \ref{algo:pics} (first three columns), and compare the results with the reconstruction of the 100\%-sampled $k$-space signal using the algorithm in \cite{Kontogiannis2021} (last column). For the reconstructed sparse signals (first three columns), we show the zero-filled velocity (rows 2,5), the reconstructed measured velocity (rows 3,6), and the reconstructed modelled velocity (rows 4,7). For the 100\%-sampled $k$-space signal (last column), we show the resulting velocity using Fourier inversion (rows 2,5), the postprocessed (unwrapped and masked) Fourier inversion (rows 3,5) and the reconstructed velocity (rows 4,7), obtained using the algorithm in \cite{Kontogiannis2021} (flow is from left to right).}
\label{fig:rec_results_1}
\end{figure*}

We first test the algorithm on sparse $k$-space signals that we generate using $\mathcal{P}_\odot$ for 5\%, 10\% and 15\% sampling. The generated sampling patterns, the reconstructed measured velocity $\bm{u}^\star$ (obtained from the phase difference), and the reconstructed modelled velocity $\bm{u}$ (N--S solution), are shown in figure \ref{fig:rec_results_1}, alongside their corresponding zero-filling solution. The fourth column of figure \ref{fig:rec_results_1} depicts the result of the image reconstruction algorithm \cite{Kontogiannis2021}, which we use when the signals are fully-sampled. For fully-sampled \mbox{$k$-space} signals, there is a one-to-one correspondence between the $k$-space and the physical space, and, consequently, we can directly reconstruct the velocity field in physical space ($\bm{u}^\star$ is known, but noisy, and fixed during the reconstruction). 

In the experiments we selected velocity encoding parameters that cause aliasing due to phase wrapping, in order to increase the SNR of the velocity images. For the 100\% sampling case with this flow, we use basic knowledge of fluid mechanics to unwrap the image by adding or subtracting multiples of $2\pi$ to aliased pixels of the phase difference image. We can do this because we already know that the velocity components have well-defined regions with fixed signs. We use this simple method of unwrapping only for the 100\% sampling case, in order to be able to reconstruct it using the algorithm in \cite{Kontogiannis2021}. For 5\%, 10\% and 15\% sampling, the present algorithm (PICS, algorithm \ref{algo:pics}) unwraps and reconstructs the measured velocity simultaneously and autonomously (figure \ref{fig:automatic_unwrapping}). This is possible with algorithm \ref{algo:pics} because it treats $\bm{u}^\star$ as a function of the unknown phases $\bm{\varphi}$, i.e. with $\bm{u}^\star$ no longer being fixed.

\begin{figure*}
\centering
\includegraphics[width=0.9\textwidth]{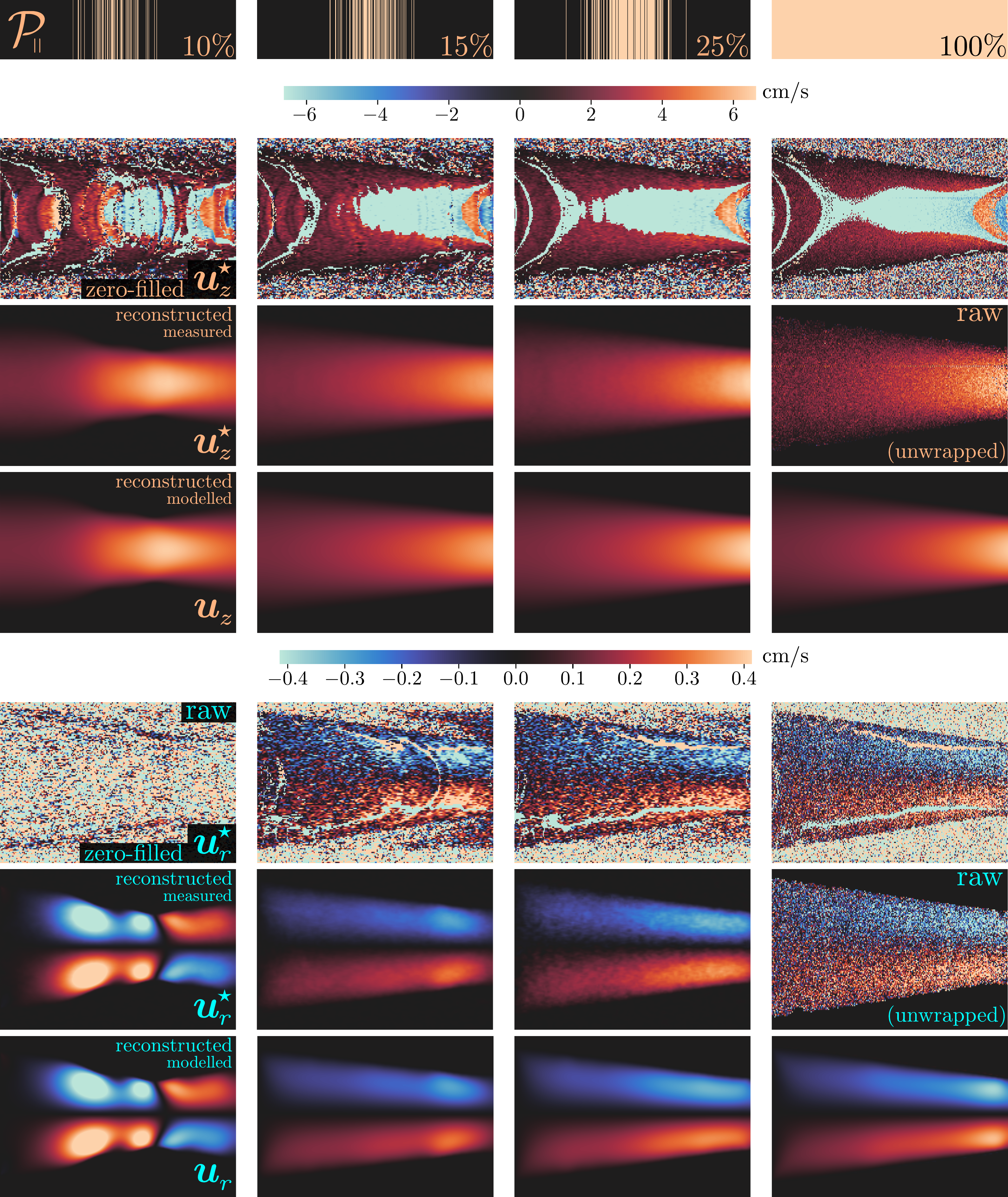}
\caption{As for figure \ref{fig:rec_results_1}, but for sparse $k$-space signals with 10\%, 15\% and 25\% $\mathcal{P}_\parallel$-sampling.}
\label{fig:rec_results_2}
\end{figure*}

\begin{figure}
\centering
\includegraphics[width=0.49\textwidth]{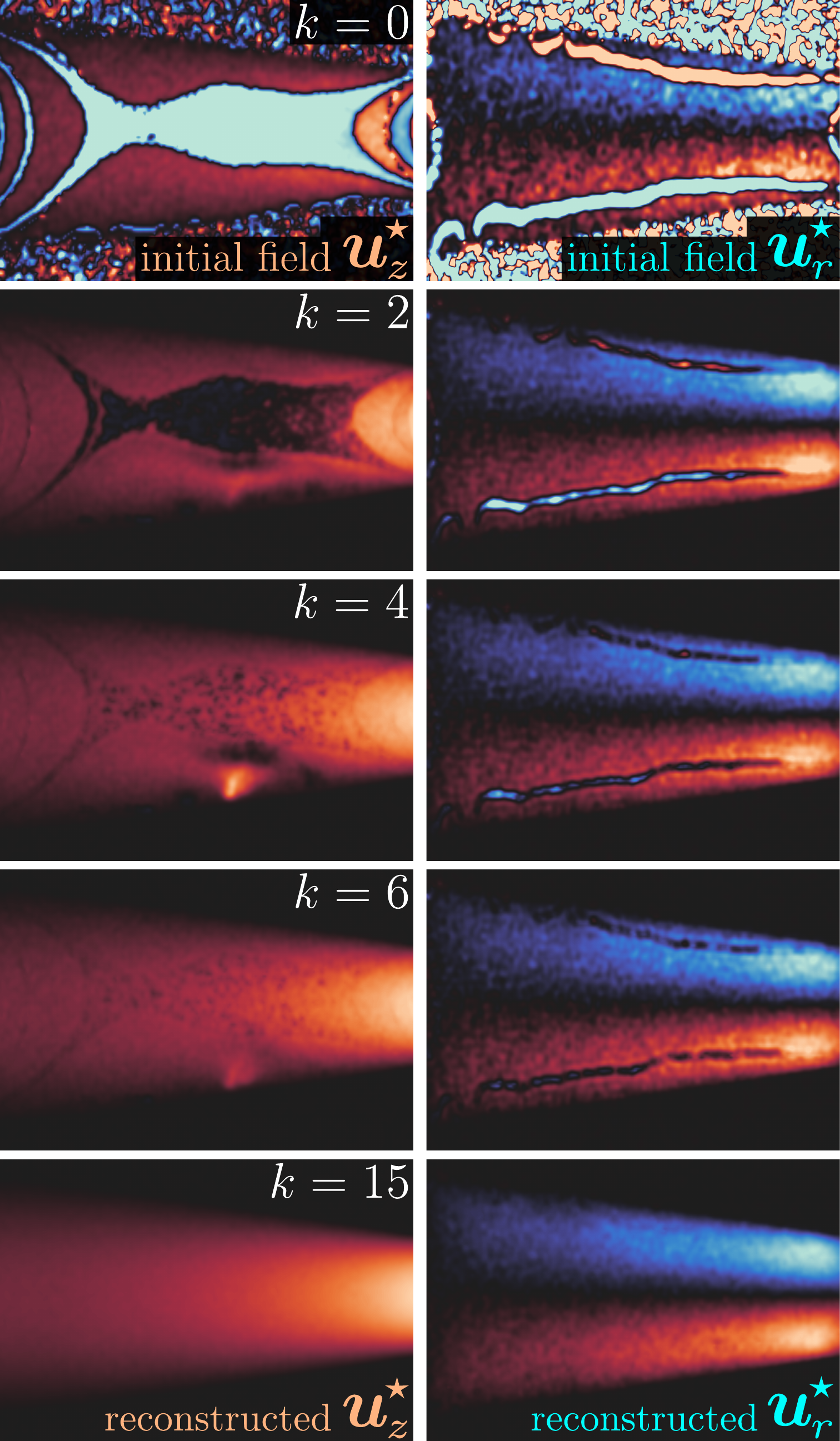}
\caption{Steps, $k$, in the iterative process that simultaneously reconstructs and unwraps the velocity. This is shown for the measured axial velocity component $u^\star_z$ (left), and the measured radial velocity component $u^\star_r$ (right), starting from their respective zero-filled solution (colormap scale shown in figure \ref{fig:rec_results_2}).}
\label{fig:automatic_unwrapping}
\end{figure}

The velocity reconstruction errors, defined by equation \eqref{eq:rec_error}, are presented in table \ref{tab:sampling_density_effect} for $\mathcal{P}_\odot$-subsampled and \mbox{$\mathcal{P}_\parallel$-subsampled} signals. We observe that for 15\% \mbox{$\mathcal{P}_\odot$-sampling}, the reconstruction errors are $\mathcal{E}^\bullet_z = 0.55$ and $\mathcal{E}^\bullet_r = 0.30$, and compare this result to the 100\%-sampling reconstruction errors of $\mathcal{E}^\bullet_z = 0.56$ and $\mathcal{E}^\bullet_r = 0.31$. The fact that the $\mathcal{P}_\odot$-subsampled signal has a lower reconstruction error than the 100\%-sampled signal may seem counterintuitive, but in reality the difference in the errors is negligible, and it can be attributed to either the technical implementation details of this large-scale problem, or in the outliers, the artefacts and the noise in the remaining 85\% of the $k$-space. For the sampling pattern $\mathcal{P}_\odot$, both the reconstruction error values (table \ref{tab:sampling_density_effect}) and a visual inspection of the velocity images (figures \ref{fig:rec_results_1},\ref{fig:gt_images}) show that 15\% sampling is sufficient to accurately reconstruct the flow through the converging nozzle. Next, we test the algorithm on sparse $k$-space signals that we generate using $\mathcal{P}_\parallel$ for 10\%, 15\% and 25\% sampling. This sampling pattern is more coherent than $\mathcal{P}_\odot$, and, therefore, it produces strong artefacts in the phase and magnitude images for low sampling densities (figure \ref{fig:rec_results_2}), making the reconstruction problem harder to solve. For the sampling pattern $\mathcal{P}_\parallel$, both the reconstruction error values (table \ref{tab:sampling_density_effect}) and a visual inspection of the velocity images (figures \ref{fig:rec_results_2},\ref{fig:gt_images}) show that 25\% sampling is sufficient in order to accurately reconstruct the flow through the converging nozzle.

Although we treat the kinematic viscosity $\nu$ as an unknown parameter, for all reconstructions we find that the posterior distribution of $\nu$ remains effectively unchanged, i.e. $\nu^\circ \simeq \mean{\nu}$ and $\sigma^\circ_\nu \simeq {\sigma}_\nu$, where $\sigma^\circ_\nu$ is the posterior standard deviation. This is because the prior information is accurate enough (see Appendix \ref{app:priors}, where the prior variance $\sigma^2_\nu$ is small) and the model $\mathscr{M}$ cannot further improve this prediction because the velocity reconstruction functional $\mathscr{U}$ is insensitive to such small changes in $\nu$ (the prior term in equation \eqref{eq:nu_grad} dominates) \cite[Section~3.5]{Kontogiannis2021}.

\subsection{Wall shear stress estimation} 
\begin{figure*}
\centering
\includegraphics[width=0.95\textwidth]{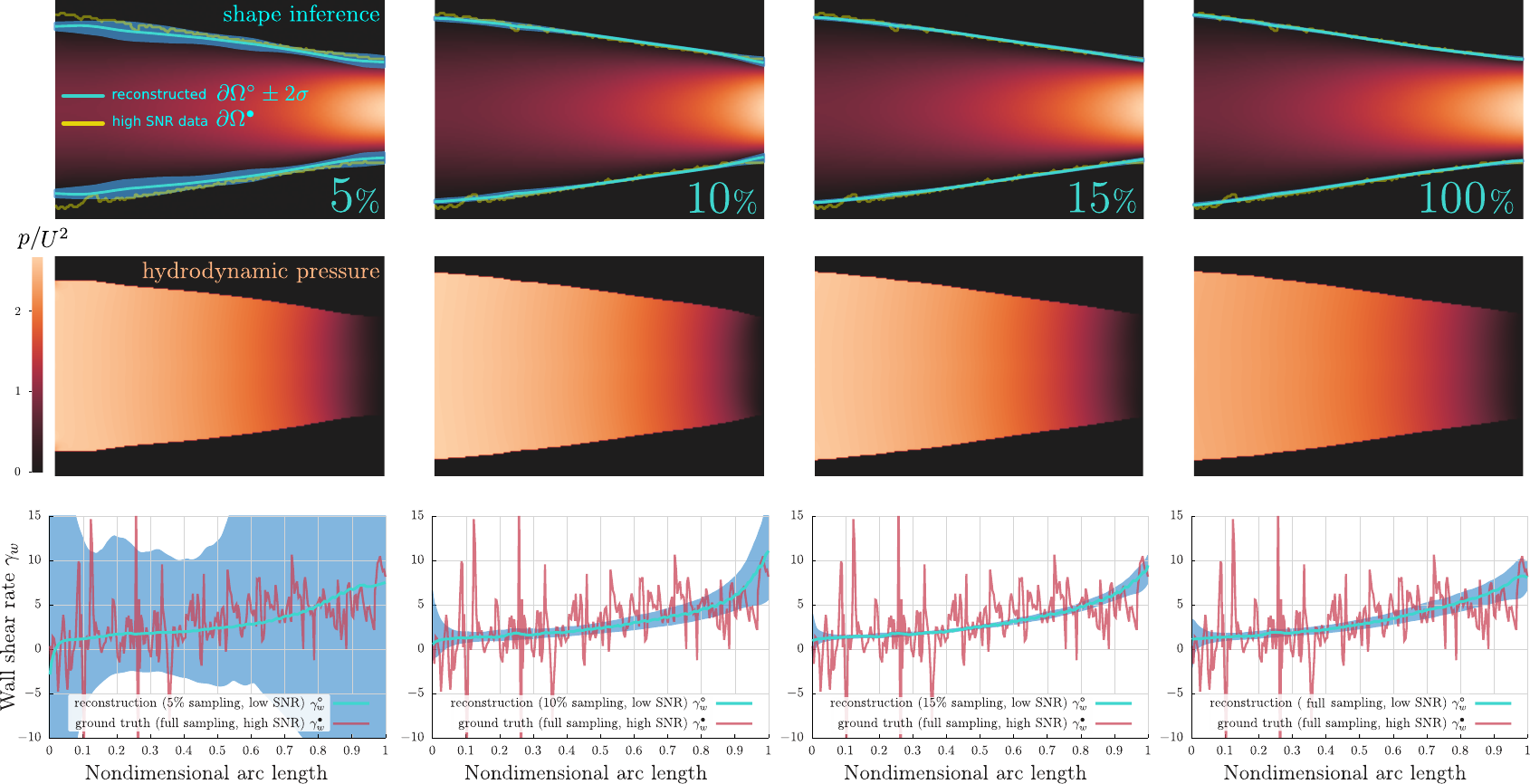}
\caption{Inferred shape, $\partial\Omega^\circ$, and velocity magnitude (first row) for 5\%, 10\% and 15\% $\mathcal{P}_\odot$-sampling using algorithm \ref{algo:pics}, and for 100\%-sampling using the algorithm in \cite{Kontogiannis2021}. The inferred shape is shown in cyan color and the 2$\sigma$ confidence interval in blue color. The yellow line denotes the ground truth of $\partial\Omega$, computed from the segmentation of the high SNR magnitude images. The inferred (nondimensional) reduced hydrodynamic pressure (second row) $p^\circ/U^2$, where $U = 4.39$ cm/s, is immediately obtained from the Navier--Stokes problem without any additional computation (see algorithm \ref{algo:pics}). The predicted wall shear rate $\gamma^\circ_w \equiv \bm{\tau}\bm{\cdot}\partial_{\bm{\nu}}\bm{u}^\circ$ (third row), where $\bm{\tau}$ is the unit tangent vector of $\partial\Omega^\circ$, is shown in cyan, and the 2$\sigma$ confidence interval, estimated by propagating the uncertainties of $\bm{x}^\circ$ through the Navier--Stokes problem, is shown in blue. The red line denotes the wall shear rate computed from the high SNR velocity data ($\gamma^\bullet_w $). The wall shear stress $\tau_w$ is found by multiplying the wall shear rate by the dynamic viscosity $\mu$ ($= \rho\nu$, where $\rho = 1183.6$ kg/m\textsuperscript{3}), i.e. $\tau_w = \mu \gamma_w$.}
\label{fig:wss_estimation}
\end{figure*}

Finally we compute the reconstructed wall shear rate $\gamma^\circ_w$ (figure \ref{fig:wss_estimation}) using the reconstructed (noiseless, $\text{SNR}=\infty$) velocity field $\bm{u}^\circ$ ($=\mathcal{Z}\bm{x}^\circ$) and the inferred shape $\partial\Omega^\circ$, and compare it with the ground truth wall shear rate $\gamma^\bullet_w$, computed for the ground truth (high SNR) velocity measurement $\bm{u}^\bullet$ (figure \ref{fig:gt_images}) on the ground truth (high SNR) shape $\partial\Omega^\bullet$. We further compute the uncertainty in $\gamma^\circ_w$ by propagating the uncertainty of the reconstructed N--S unknowns $\bm{x}^\circ$ through the N--S problem. We observe that the ground truth wall shear rate is particularly noisy because PC-MRI suffers from low resolution and partial volume effects \cite{Bouillot2018,Saito2020} near the boundaries $\partial\Omega$. Using standard image processing algorithms, it is possible to either denoise $\gamma^\bullet_w$ directly, or compute $\gamma^\bullet_w$ for a denoised velocity $\bm{u}^\bullet$ and a smoothed boundary $\partial\Omega^\bullet$. This denoised velocity field will not, however, be consistent with the new smoothed boundary (the no-slip wall boundary condition will not be satisfied), and the steep, near-wall velocity gradients that dictate the wall shear rate magnitude will not be accurately recovered. On the other hand, our inverse Navier--Stokes boundary value problem (algorithm \ref{algo:pics}), jointly reconstructs the velocity field and infers the boundary. This accurately reconstructs the near-wall velocity distribution, and, if needed, can be made more accurate by increasing the N--S model resolution. This method infers the boundary by drawing information from both the velocity field and the spin density (magnitude) images so that geometric errors that affect the estimated wall shear stress distribution can be minimized.       

\subsection{Further reducing sampling density}

\subsubsection{Targeted sampling for the N--S unknowns $\bm{x}$}
At this point, it is important to mention that the sampling density can be further decreased (for fixed reconstruction error) if the $\textit{a priori}$ information on $\bm{x}$ becomes more accurate. The present method indicates that when the Navier--Stokes problem is an accurate and descriptive model, only the boundary $\partial\Omega$, the boundary conditions $\bm{g}_i,\bm{g}_o$, and the kinematic viscosity are needed in order to find the velocity field (see figure \ref{fig:sparsity_extended}). Therefore, to reduce signal acquisition time further, we can obtain more accurate spin density (magnitude) images separately (without encoding velocity) and use them as priors for $\partial\Omega$ ($\bar{\phi}_\pm$). Similarly, we can obtain $d-1$ dimensional scans for the inlet velocity boundary condition, and use them as priors for $\bar{\bm{g}}_i$. Conventional MRI or PC-MRI cannot measure $\bm{g}_o$ and $\nu$, but $\nu$ is usually known with high certainty compared to the other parameters. Targeted sampling using a Navier--Stokes problem is an interesting generalization \cite{Duarte2011} of sparsity in CS, and closely follows the concept of compressed sampling. 

\subsubsection{Design of optimal sampling patterns $\mathcal{P}$}
Another way to further decrease the sampling density is to design optimal sampling patterns \cite{Gladden2017}, also taking into account the constraints of the PC-MRI signal acquisition protocols. For example, in section \ref{sec:rec_results} we found that the generated 15\% $\mathcal{P}_\odot$-sampling pattern produces slightly better results than the generated 25\% \mbox{$\mathcal{P}_\parallel$-sampling} pattern, but note that the former adequately covers the center of $k$-space (figure \ref{fig:rec_results_1}), while the latter leaves gaps (figure \ref{fig:rec_results_2}). This can be corrected by fully-sampling the central region of $k$-space and then sparsely-sampling regions that are not as important. To identify the important regions, we can use our algorithm to backpropagate the errors from the velocity images to the $k$-space. The physics-informed design of optimal sparse sampling patterns is left for future work.

{\color{black}
\subsection{Extension to multi-coil parallel MRI}
Our algorithm extends to parallel MRI \cite{Deshmane2012} using a SENSE-type approach \cite{Pruessmann1999}, in which the individual coil sensitivities are inferred. In brief, the $k$-space reconstruction error, which in non-parallel MRI is given by \eqref{eq:kspace_rec_consistency}, is now given by
\begin{align}
\mathscr{E}(\bm{\varphi},\bm{\rho},\bm{\lambda}) \equiv \sum^{4d}_{j=1}\frac{1}{2\sigma_j^2}\sum_{k=1}^{\Lambda}\norm{s^\star_{jk}-\mathcal{P}\mathcal{F}\lambda_k(\rho_je^{i\varphi_j})}^2_{L^2(I_s)}\quad,
\label{eq:k_space_rec_multicoil}
\end{align}
where $\Lambda$ is the total number of coils, and $\lambda_k \in L^2(I)$ is the (unknown) sensitivity of the $k$-th coil. 

To infer $\bm{\lambda}$ we introduce an additional regularization term
\begin{gather}
\mathscr{R}_{\lambda} \equiv \frac{1}{2}\sum^{\Lambda}_{k=1}\norm{\lambda_k - \overline{\lambda}_k}^2_{\mathcal{C}_{\lambda_k}(I)}\quad,
\label{eq:coil_sensitivities}
\end{gather}
where $\overline{\lambda}_k$ is the prior mean (or initial guess), and $\mathcal{C}_{\lambda_k}$ is a user-selected covariance operator. Since coil sensitivity fields are often assumed to be smooth, a reasonable choice is an operator that is similar to the Bessel potential \cite[Chapter~V.3]{stein1970}
\begin{gather}
\mathcal{C}_{\lambda_k} = \sigma^2_{\lambda_k}\Big(\mathrm{I} - \ell^2{\Delta}\Big)^{-s}\quad,
\end{gather}
where $\sigma^2_{\lambda_k} \in \R$ is proportional to the variance of the $k$-th coil sensitivity (level of confidence in the prior), $\ell \in \R$ is a characteristic length scale (length scales smaller than $\ell$ are suppressed), and $\R \ni s > 0$ is a user-selected parameter based on smoothness assumptions ($s \gtrsim 1$). {\color{black}It is worth noting that the coil sensitivity priors $\overline{\lambda}_k$ can be obtained with a method such as ESPIRiT \cite{espirit2014}, as in \cite{Sun2017}.}

Replacing functional \eqref{eq:kspace_rec_consistency} with \eqref{eq:k_space_rec_multicoil}, and adding \eqref{eq:coil_sensitivities} to the Lagrangian $\mathscr{J}$, which is given by \eqref{eq:lagrangian}, the corresponding gradients $D_{\lambda_k}\mathscr{J}$ and the steepest ascent directions $\widehat{D}_{\lambda_k}\mathscr{J}$ can be derived. Algorithm \ref{algo:pics} must be augmented to include one more reconstruction step, during which the steepest ascent directions $\widehat{D}_{\bm{\lambda}}\mathscr{J}$ and the step size $\tau$ are first computed, and the coil sensitivities are then updated by
\begin{equation}
\bm{\lambda}_{k+1} \leftarrow \bm{\lambda}_k - \tau \widehat{D}_{\bm{\lambda}}\mathscr{J}\quad,
\end{equation} 
where $k$ is the iteration index of algorithm \ref{algo:pics}. In this way we minimize \mbox{$\mathscr{J}({\color{gray}\bm{\varphi}_{k+1}},{\color{gray}\bm{x}_{k+1}},{\color{gray}\bm{\rho}_{k+1}},{\color{gray}\bm{c}_{k+1}},\bm{\lambda})$} with respect to the coil sensitivities $\bm{\lambda}$, with all other unknowns being fixed.
}

\section{Conclusion}
We propose a physics-informed compressed sensing (PICS) reconstruction method that optimally combines computational fluid dynamics with phase-contrast magnetic resonance imaging. By adopting a Bayesian framework, we use \textit{a priori} knowledge to regularize the inverse problem and find the most likely measured velocity field, segmentation, and \mbox{Navier--Stokes} twin (maximum \textit{a posteriori} estimates) for the flow through a converging nozzle. We show that the physics-informed reconstruction of the sparse (15\% $\mathcal{P}_\odot$-sampling), low SNR signals, compares well to fully-sampled, high SNR images, of the same flow.

PICS extends the capabilities of conventional compressed sensing methods for phase-contrast MRI. In particular, it integrates an optimal filter (the regularity of the velocity field is controled by the Navier--Stokes problem), it infers the hydrodynamic pressure without any additional computation, and allows us to estimate the wall shear stresses at negligible additional numerical cost. At the same time it provides us with a digital twin, which can subsequently be used to simulate different flow conditions, and paves the way for the development of new applications in patient-specific cardiovascular modeling. Our formulation naturally extends to three-dimensional and periodic flows in complicated geometries, and the \mbox{Navier--Stokes} problem can be updated to model complex (e.g. non-Newtonian) fluid flows. \revall{We are currently developing efficient numerical algorithms to implement our formulation in three dimensions.}  

\section*{Acknowledgments}
The authors would like to thank Scott V. Elgersma and Andrew J. Sederman for generating the PC-MRI dataset, which is the same as that of \cite{Kontogiannis2021}, and for their useful input and comments during the preparation of this manuscript. Author A.K. is financially supported by the W.D. Armstrong scholarship from the Cambridge Trusts.

{\appendices
\section{E--L equations for $\bm{\varphi},\bm{\rho},\bm{c}$ and $\bm{x}$}
\label{app:sadirections}
Here we expand \eqref{eq:el_phase}-\eqref{eq:el_segm} in order to obtain explicit relations for the corresponding generalized gradients ${D}_{(\cdot)}\mathscr{J}$, which are defined by \eqref{eq:first_variation}. Note that the steepest ascent directions, which are defined by \eqref{eq:sadirection}, are given by
\begin{equation}
\widehat{D}_{(\cdot)}\mathscr{J} = \mathcal{C}_{(\cdot)}{D}_{(\cdot)}\mathscr{J}\quad.
\end{equation}

For the phases and magnitudes we find
\begin{align}
{D}_{{\varphi}_j}&\mathscr{J} = \chi_{I_w}(-1)^{j\textrm{div}2}c_k~\mathcal{C}_{u_k}^{-1}\big(u^\star_k - \mathcal{S}u_k\big) \nonumber\\ &+ \mathcal{C}^{-1}_{\varphi_j}\Re\Big(ie^{i\varphi_j}\big(e^{i\varphi_j}-e^{i\bar{\varphi}_j}\big)^\dagger\Big) \nonumber\\
&+ \sigma_j^{-2}\rho_j~\Im\Big(e^{i\varphi_j}\big(\mathcal{F}^{-1}\mathcal{P}^\dagger(s^\star_j-\mathcal{P}\mathcal{F}\rho_j e^{i\varphi_j})\big)^\dagger\Big)\ ,\label{eq:phase_grad}\\
{D}_{{\rho}_j}&\mathscr{J} = \chi_{I_w}(\sqrt{4d}\ \sigma_{\rho_j})^{-2}\Big(\big(\rho_j-\alpha\big)\mathcal{S}\mathcal{H}(\sdist) \nonumber\\ 
&+\big(\rho_j-\beta\big)\big(\mathcal{S}\mathcal{H}(\sdist)-1\big)\Big) + \mathcal{C}^{-1}_{\rho_j}\big(\rho_j-\bar{\rho}_j) \nonumber \\ &-\sigma_j^{-2}~\Re\Big(e^{i\varphi_j}\big(\mathcal{F}^{-1}\mathcal{P}^\dagger(s^\star_j-\mathcal{P}\mathcal{F}\rho_j e^{i\varphi_j})\big)^\dagger\Big)\ ,\label{eq:mag_grad}
\end{align}
for $j=1,\dots,4d$, where $k\equiv(j-1)\textrm{div}4 +1$, $\chi_{I_w}$ is the characteristic function of $I_w \subseteq I$. Note that \mbox{${D}_{{\bm{\varphi}}}\mathscr{J}(\bm{\varphi},\bm{u},\bm{\rho}) \in L^2(I)$}, and ${D}_{{\bm{\rho}}}\mathscr{J}(\bm{\rho},\sdist,\bm{\varphi},\alpha,\beta) \in L^2(I)$. 

For the segmentation constants we find
\begin{align} 
{D}_{\alpha}\mathscr{J} &= -\sum_{j=1}^{4d} (\sqrt{4d}\ \sigma_{\rho_j})^{-2}\int_\Omega \rho_j - \alpha\quad, \label{eq:alpha_grad} \\
{D}_{\beta}\mathscr{J} &=  -\sum_{j=1}^{4d} (\sqrt{4d}\ \sigma_{\rho_j})^{-2}\int_{I_w-\Omega} \rho_j - \beta\quad \label{eq:beta_grad},
\end{align}
where $I_w-\Omega$ denotes the complement of $\Omega$ in $I_w$. Note that {${D}_{{\alpha}}\mathscr{J}(\bm{\rho},\sdist,\alpha) \in \R$}, and ${D}_{\beta}\mathscr{J}(\bm{\rho},\sdist,\beta) \in \R$. Fixing $\bm{\rho}$ and $\Omega$ (or, equivalently, $\sdist$), and setting \eqref{eq:alpha_grad} and \eqref{eq:beta_grad} to zero, we find the following explicit relations for the constants $\alpha,\beta$ that minimize $\mathscr{J}$ 
\begin{align}
\alpha &= \sum_{j=1}^{4d} (\sqrt{4d\abs{\Omega}}\sigma_{\rho_j})^{-2} \int_\Omega \rho_j\quad, \label{eq:alpha_value}\\
\beta &= \sum_{j=1}^{4d} (\sqrt{4d\abs{I_w-\Omega}}\sigma_{\rho_j})^{-2} \int_{I_w-\Omega} \rho_j\quad. \label{eq:beta_value}
\end{align}

The gradients for the N--S unknowns $\bm{x}$ are drawn directly from \cite[Sections~2.3,2.4]{Kontogiannis2021}
\begin{align}
{D}_{\bm{g}_i}\mathscr{J} &= \nu\partial_{\bm{\nu}}\bm{v}-q\bm{\nu}- \eta\bm{v} + \invcgi\big(\bm{g}_i-{\mean{\bm{g}}_i} \big)\ , \label{eq:gi_grad}\\ 
{D}_{\bm{g}_o}\mathscr{J} &= \bm{v} + \invcgo\big(\bm{g}_o-\mean{\bm{g}}_o\big)\ ,\label{eq:go_grad}\\
{D}_{\nu}\mathscr{J} &= \nabla\bm{v}\bm{:}\nabla\bm{u} + \sigma_\nu^{-2} \big(\nu-\mean{\nu}\big)\ ,\label{eq:nu_grad}\\
{D}_{\sdist}\mathscr{(J-\mathscr{S})} &= \zetaext + \mathcal{C}^{-1}_\sdist\big(\mean{\phi}_\pm-\sdist \big)\label{eq:sdist_grad}\ ,
\end{align}
where $\zetaext \in L^2(M)$ is an extension of the shape gradient \mbox{$\zeta \in L^2(\Gamma)$} that is generated by an advection-diffusion problem \cite[Section~2.4]{Kontogiannis2021}, and the shape gradient is given by
\begin{align}
\zeta = \partial_{\bm{\nu}}\bm{u}\bm{\cdot}\big(-\nu\partial_{\bm{\nu}}\bm{v}+q\bm{\nu}\big) \quad .
\end{align}
The segmentation functional $\mathscr{S}$, given by \eqref{eq:segm_functional}, was not included in \cite{Kontogiannis2021} because the boundary $\partial\Omega$ was inferred only from the velocity images. Here, $\mathscr{S}$ allows us to draw information from the spin density (magnitude) images in order to better infer the boundary. For example, near regions of low velocity gradients (e.g. nearly separated flow) the velocity field alone is not informative. Consequently, taking into account $\mathscr{S}$, we add the following term in \eqref{eq:sdist_grad} 
\begin{align}
D_{\sdist}\mathscr{S} = \zetaext_\mathscr{S} \in L^2(M) \quad,
\end{align}
where $\zetaext_\mathscr{S}$ is an advection-diffusion extension of $\zeta_\mathscr{S} \in L^2(\Gamma)$, similar to $\zeta^\circ$, and
\begin{align}
\zeta_\mathscr{S} = \delta_\Gamma \sum_{j=1}^{4d} (\sqrt{8d}\sigma_j)^{-2} \mathcal{S}^\dagger\Big((\rho_j-\alpha)^2-(\rho_j-\beta)^2\Big), \label{eq:geomflow_segm}
\end{align}
where $\delta_\Gamma$ is the Dirac measure on $\Gamma$, or, equivalently, the characteristic function of $\Gamma$ ($\delta_\Gamma = \chi_\Gamma$).

\section{Phase-Contrast MRI experiment}
\label{app:pcmri_exp}
The PC-MRI experiment for the flow through the converging nozzle is described in detail in \cite[Section~3.4]{Kontogiannis2021}. In this paper, in order to demonstrate the de-aliasing capability of algorithm \ref{algo:pics}, we use $k$-space signals (of the same dataset) that produced severely aliased velocity images. These signals correspond to slightly different flow conditions than those described in \cite{Kontogiannis2021}. They were acquired for a 70 wt\%, instead of a 40 wt\%, glycerol in water solution, and the Reynolds number was 22.4. We found the $T_1$ relaxation time of this glycerol solution to be $313$ms. For the high SNR images, the repetition time was $0.5$s, which resulted in a total acquisition time of 68 minutes per velocity image. For the low SNR images, the repetition time was $150$ms, which resulted in a total acquisition time of 2.6 minutes per velocity image. The total acquisition times correspond to fully-sampled $k$-spaces. 

\section{Prior information for the N--S unknowns}
The prior variances of the Navier--Stokes unknowns are given in table \ref{tab:ns_priors_confidence}. Note that, for 5\% $\mathcal{P}_\odot$/$\mathcal{P}_\parallel$-sampling, instead of $\sigma_\sdist = 2h_z$ we set ${\sigma_\sdist=10h_z}$ as the prior information on the geometry becomes less accurate due to subsampling artefacts. \revall{For a more detailed discussion of the regularization parameters, see \cite[Section~3.6]{Kontogiannis2021}.}
\label{app:priors}
\begin{table}[!h]
\centering
  \caption{Confidence level of prior information.}
  \label{tab:ns_priors_confidence}
  \begin{tabular}{lcccc}
        & shape $\partial\Omega$& inlet $\bm{g}_i$& viscosity $\nu$ & resolution\\
        $\mathcal{P}$ & $(\sigma_\sdist,\text{Re}_{\sdist},\text{Re}_{\zeta})$ & $(\sigma_{\bm{g}_i}$[cm/s], $\ell)$ & $\sigma_\nu$ [m$^2$/s] & $h_z\times h_r$[$\mu$m]\\[2.5pt]
        \hline\\ 
        $\mathcal{P}_\odot$     & (2$h_z$, 0.025, 0.025) & (1.0, 3$h_r$) & 2.54$\times$10$^6$ & 165$\times$223\\[3pt]
        $\mathcal{P}_\parallel$ & (2$h_z$, 0.010, 0.010) & (1.0, 3$h_r$) & 2.54$\times$10$^6$ & 165$\times$223
  \end{tabular}
\end{table}
}

\bibliographystyle{IEEEtran}
\bibliography{main.bib}

\begin{thebibliography}{10}
\providecommand{\url}[1]{#1}
\csname url@samestyle\endcsname
\providecommand{\newblock}{\relax}
\providecommand{\bibinfo}[2]{#2}
\providecommand{\BIBentrySTDinterwordspacing}{\spaceskip=0pt\relax}
\providecommand{\BIBentryALTinterwordstretchfactor}{4}
\providecommand{\BIBentryALTinterwordspacing}{\spaceskip=\fontdimen2\font plus
\BIBentryALTinterwordstretchfactor\fontdimen3\font minus
  \fontdimen4\font\relax}
\providecommand{\BIBforeignlanguage}[2]{{%
\expandafter\ifx\csname l@#1\endcsname\relax
\typeout{** WARNING: IEEEtran.bst: No hyphenation pattern has been}%
\typeout{** loaded for the language `#1'. Using the pattern for}%
\typeout{** the default language instead.}%
\else
\language=\csname l@#1\endcsname
\fi
#2}}
\providecommand{\BIBdecl}{\relax}
\BIBdecl

\bibitem{Gudbjartsson1995}
H.~Gudbjartsson and S.~Patz, ``{The Rician Distribution of Noisy MRI Data},''
  \emph{Magnetic Resonance in Medicine}, vol.~34, no.~6, pp. 910--914, 1995.

\bibitem{Chang2000}
S.~G. Chang, B.~Yu, and M.~Vetterli, ``{Adaptive wavelet thresholding for image
  denoising and compression},'' \emph{IEEE Transactions on Image Processing},
  vol.~9, no.~9, pp. 1532--1546, 2000.

\bibitem{Pascal2012}
G.~Pascal, ``{Rudin--Osher--Fatemi Total Variation Denoising using Split
  Bregman},'' \emph{Image Processing On Line}, vol.~2, no.~1, pp. 74--95, 2012.

\bibitem{Fatouraee2003}
N.~Fatouraee and A.~A. Amini, ``{Regularization of flow streamlines in
  multislice phase-contrast MR imaging},'' \emph{IEEE Transactions on Medical
  Imaging}, vol.~22, no.~6, pp. 699--709, 2003.

\bibitem{Ong2015}
F.~Ong, M.~Uecker, U.~Tariq, A.~Hsiao, M.~T. Alley, S.~S. Vasanawala, and
  M.~Lustig, ``{Robust 4D flow denoising using divergence-free wavelet
  transform},'' \emph{Magnetic Resonance in Medicine}, vol.~73, no.~2, pp.
  828--842, 2015.

\bibitem{Mura2016}
J.~Mura, A.~M. Pino, J.~Sotelo, I.~Valverde, C.~Tejos, M.~E. Andia,
  P.~Irarr{\'{a}}zaval, and S.~Uribe, ``{Enhancing the Velocity Data From 4D
  Flow MR Images by Reducing its Divergence},'' \emph{IEEE Transactions on
  Medical Imaging}, vol.~35, no.~10, pp. 2353--2364, 2016.

\bibitem{Koltukluoglu2018}
T.~S. Koltukluo{\u{g}}lu and P.~J. Blanco, ``{Boundary control in computational
  haemodynamics},'' \emph{Journal of Fluid Mechanics}, vol. 847, pp. 329--364,
  2018.

\bibitem{Funke2019}
S.~W. Funke, M.~Nordaas, {\O}.~Evju, M.~S. Aln{\ae}s, and K.~A. Mardal,
  ``{Variational data assimilation for transient blood flow simulations:
  Cerebral aneurysms as an illustrative example},'' \emph{International Journal
  for Numerical Methods in Biomedical Engineering}, vol.~35, no.~1, pp. 1--27,
  2019.

\bibitem{Kontogiannis2021}
A.~Kontogiannis, S.~V. Elgersma, A.~J. Sederman, and M.~P. Juniper, ``{Joint
  reconstruction and segmentation of noisy velocity images as an inverse
  Navier--Stokes problem},'' \emph{Journal of Fluid Mechanics}, vol. 944, p.
  A40, 2022.

\bibitem{Candes2006}
E.~J. Cand{\`{e}}s, J.~Romberg, and T.~Tao, ``{Robust Uncertainty Principles :
  Exact Signal Frequency Information},'' \emph{IEEE Transactions on Information
  Theory}, vol.~52, no.~2, pp. 489--509, 2006.

\bibitem{Donoho2006}
D.~L. Donoho, ``{Compressed sensing},'' \emph{IEEE Transactions on Information
  Theory}, vol.~52, no.~4, pp. 1289--1306, 2006.

\bibitem{Lustig2007}
M.~Lustig, D.~Donoho, and J.~M. Pauly, ``{Sparse MRI: The application of
  compressed sensing for rapid MR imaging},'' \emph{Magnetic Resonance in
  Medicine}, vol.~58, no.~6, pp. 1182--1195, 2007.

\bibitem{Benning2014}
\BIBentryALTinterwordspacing
M.~Benning, L.~Gladden, D.~Holland, C.~B. Sch{\"{o}}nlieb, and T.~Valkonen,
  ``{Phase reconstruction from velocity-encoded MRI measurements - A survey of
  sparsity-promoting variational approaches},'' \emph{Journal of Magnetic
  Resonance}, vol. 238, pp. 26--43, 2014. [Online]. Available:
  \url{http://dx.doi.org/10.1016/j.jmr.2013.10.003}
\BIBentrySTDinterwordspacing

\bibitem{Holland2010}
\BIBentryALTinterwordspacing
D.~J. Holland, D.~M. Malioutov, A.~Blake, A.~J. Sederman, and L.~F. Gladden,
  ``{Reducing data acquisition times in phase-encoded velocity imaging using
  compressed sensing},'' \emph{Journal of Magnetic Resonance}, vol. 203, no.~2,
  pp. 236--246, 2010. [Online]. Available:
  \url{http://dx.doi.org/10.1016/j.jmr.2010.01.001}
\BIBentrySTDinterwordspacing

\bibitem{Roberts2013}
T.~Roberts, N.~Kingsbury, and D.~J. Holland, ``{Sparse recovery of complex
  phase-encoded velocity images using iterative thresholding},'' \emph{2013
  IEEE International Conference on Image Processing, ICIP 2013 - Proceedings},
  pp. 350--354, 2013.

\bibitem{Zhao2012}
F.~Zhao, D.~C. Noll, J.~F. Nielsen, and J.~A. Fessler, ``{Separate magnitude
  and phase regularization via compressed sensing},'' \emph{IEEE Transactions
  on Medical Imaging}, vol.~31, no.~9, pp. 1713--1723, 2012.

\bibitem{Reci2019}
A.~Reci, ``Signal sampling and processing in magnetic resonance applications,''
  Ph.D. dissertation, University of Cambridge, 2019.

\bibitem{Bredies2010}
K.~Bredies, K.~Kunisch, and T.~Pock, ``{Total generalized variation},''
  \emph{SIAM Journal on Imaging Sciences}, vol.~3, no.~3, pp. 492--526, 2010.

\bibitem{Corona2021}
\BIBentryALTinterwordspacing
V.~Corona, M.~Benning, L.~F. Gladden, A.~Reci, A.~J. Sederman, and C.-B.
  Sch{\"{o}}nlieb, ``{Joint Phase Reconstruction and Magnitude Segmentation
  from Velocity-Encoded MRI Data},'' \emph{Time-dependent Problems in Imaging
  and Parameter Identification}, pp. 1--24, 2021. [Online]. Available:
  \url{http://arxiv.org/abs/1908.05285}
\BIBentrySTDinterwordspacing

\bibitem{Kollmeier2022}
\BIBentryALTinterwordspacing
J.~M. Kollmeier, O.~Kalentev, J.~Klosowski, D.~Voit, and J.~Frahm, ``{Velocity
  vector reconstruction for real-time phase-contrast MRI with radial Maxwell
  correction},'' \emph{Magnetic Resonance in Medicine}, vol.~87, no.~4, pp.
  1863--1875, 2022. [Online]. Available:
  \url{https://onlinelibrary.wiley.com/doi/abs/10.1002/mrm.29108}
\BIBentrySTDinterwordspacing

\bibitem{Bakhshinejad2017}
\BIBentryALTinterwordspacing
A.~Bakhshinejad, A.~Baghaie, A.~Vali, D.~Saloner, V.~L. Rayz, and R.~M.
  D’Souza, ``{Merging computational fluid dynamics and 4D Flow MRI using
  proper orthogonal decomposition and ridge regression},'' \emph{Journal of
  Biomechanics}, vol.~58, pp. 162--173, 2017. [Online]. Available:
  \url{https://www.sciencedirect.com/science/article/pii/S0021929017302531}
\BIBentrySTDinterwordspacing

\bibitem{Toger2020}
\BIBentryALTinterwordspacing
J.~Töger, M.~J. Zahr, N.~Aristokleous, K.~Markenroth~Bloch, M.~Carlsson, and
  P.-O. Persson, ``{Blood flow imaging by optimal matching of computational
  fluid dynamics to 4D-flow data},'' \emph{Magnetic Resonance in Medicine},
  vol.~84, no.~4, pp. 2231--2245, 2020. [Online]. Available:
  \url{https://onlinelibrary.wiley.com/doi/abs/10.1002/mrm.28269}
\BIBentrySTDinterwordspacing

\bibitem{Sotelo2016}
J.~Sotelo, J.~Urbina, I.~Valverde, C.~Tejos, P.~Irarrazaval, M.~E. Andia,
  S.~Uribe, and D.~E. Hurtado, ``{3D Quantification of Wall Shear Stress and
  Oscillatory Shear Index Using a Finite-Element Method in 3D CINE PC-MRI Data
  of the Thoracic Aorta},'' \emph{IEEE Transactions on Medical Imaging},
  vol.~35, no.~6, pp. 1475--1487, 2016.

\bibitem{Zhang2020}
J.~Zhang, M.~C. Brindise, S.~Rothenberger, S.~Schnell, M.~Markl, D.~Saloner,
  V.~L. Rayz, and P.~P. Vlachos, ``{4D Flow MRI Pressure Estimation Using
  Velocity Measurement-Error-Based Weighted Least-Squares},'' \emph{IEEE
  Transactions on Medical Imaging}, vol.~39, no.~5, pp. 1668--1680, 2020.

\bibitem{Berg2014}
P.~Berg, D.~Stucht, G.~Janiga, O.~Beuing, O.~Speck, and D.~Th{\'{e}}venin,
  ``{Cerebral blood flow in a healthy circle of willis and two intracranial
  aneurysms: Computational fluid dynamics versus four-dimensional
  phase-contrast magnetic resonance imaging},'' \emph{Journal of Biomechanical
  Engineering}, vol. 136, no.~4, pp. 1--9, 2014.

\bibitem{Id2019}
J.~Id and P.~Berg, ``{2018 ( MATCH )— Phase Ib : Effect of morphology on
  hemodynamics},'' vol. 2018, 2019.

\bibitem{Duarte2011}
M.~F. Duarte and Y.~C. Eldar, ``{Structured compressed sensing: From theory to
  applications},'' \emph{IEEE Transactions on Signal Processing}, vol.~59,
  no.~9, pp. 4053--4085, 2011.

\bibitem{Bright2013}
I.~Bright, G.~Lin, and J.~N. Kutz, ``{Compressive sensing based machine
  learning strategy for characterizing the flow around a cylinder with limited
  pressure measurements},'' \emph{Physics of Fluids}, vol.~25, no.~12, 2013.

\bibitem{Ferdian2020}
\BIBentryALTinterwordspacing
E.~Ferdian, A.~Suinesiaputra, D.~J. Dubowitz, D.~Zhao, A.~Wang, B.~Cowan, and
  A.~A. Young, ``{4DFlowNet: Super-Resolution 4D Flow MRI Using Deep Learning
  and Computational Fluid Dynamics},'' \emph{Frontiers in Physics}, vol.~8,
  2020. [Online]. Available:
  \url{https://www.frontiersin.org/articles/10.3389/fphy.2020.00138}
\BIBentrySTDinterwordspacing

\bibitem{Rutkowski2021}
D.~R. Rutkowski, A.~Roldán-Alzate, and K.~M. Johnson, ``{Enhancement of
  cerebrovascular 4D flow MRI velocity fields using machine learning and
  computational fluid dynamics simulation data},'' \emph{Scientific Reports},
  vol.~11, no. 10240, 2021.

\bibitem{9420272}
M.~J. Muckley, B.~Riemenschneider, A.~Radmanesh, S.~Kim, G.~Jeong, J.~Ko,
  Y.~Jun, H.~Shin, D.~Hwang, M.~Mostapha, S.~Arberet, D.~Nickel, Z.~Ramzi,
  P.~Ciuciu, J.-L. Starck, J.~Teuwen, D.~Karkalousos, C.~Zhang, A.~Sriram,
  Z.~Huang, N.~Yakubova, Y.~W. Lui, and F.~Knoll, ``{Results of the 2020
  fastMRI Challenge for Machine Learning MR Image Reconstruction},'' \emph{IEEE
  Transactions on Medical Imaging}, vol.~40, no.~9, pp. 2306--2317, 2021.

\bibitem{doi:10.1073/pnas.2107151119}
\BIBentryALTinterwordspacing
M.~J. Colbrook, V.~Antun, and A.~C. Hansen, ``{The difficulty of computing
  stable and accurate neural networks: On the barriers of deep learning and
  Smale's 18th problem},'' \emph{Proceedings of the National Academy of
  Sciences}, vol. 119, no.~12, p. e2107151119, 2022. [Online]. Available:
  \url{https://www.pnas.org/doi/abs/10.1073/pnas.2107151119}
\BIBentrySTDinterwordspacing

\bibitem{Bungert_2020}
\BIBentryALTinterwordspacing
L.~Bungert, M.~Burger, Y.~Korolev, and C.-B. Schönlieb, ``{Variational
  regularisation for inverse problems with imperfect forward operators and
  general noise models},'' \emph{Inverse Problems}, vol.~36, no.~12, p. 125014,
  dec 2020. [Online]. Available: \url{https://doi.org/10.1088/1361-6420/abc531}
\BIBentrySTDinterwordspacing

\bibitem{doi:10.1137/20M1338460}
\BIBentryALTinterwordspacing
S.~Lunz, A.~Hauptmann, T.~Tarvainen, C.-B. Sch\"{o}nlieb, and S.~Arridge, ``{On
  Learned Operator Correction in Inverse Problems},'' \emph{SIAM Journal on
  Imaging Sciences}, vol.~14, no.~1, pp. 92--127, 2021. [Online]. Available:
  \url{https://doi.org/10.1137/20M1338460}
\BIBentrySTDinterwordspacing

\bibitem{Chan2001}
T.~F. Chan and L.~A. Vese, ``{Active contours without edges},'' \emph{IEEE
  Transactions on Image Processing}, vol.~10, no.~2, pp. 266--277, 2001.

\bibitem{Getreuer2012}
P.~Getreuer, ``{Chan--Vese Segmentation},'' \emph{Image Processing On Line},
  vol.~2, pp. 214--224, 2012.

\bibitem{Nitsche1971}
J.~Nitsche, ``{{\"{U}}ber ein Variationsprinzip zur L{\"{o}}sung von
  Dirichlet-Problemen bei Verwendung von Teilr{\"{a}}umen, die keinen
  Randbedingungen unterworfen sind},'' \emph{Abhandlungen aus dem
  Mathematischen Seminar der Universit{\"{a}}t Hamburg}, vol.~36, no.~1, pp.
  9--15, 1971.

\bibitem{Fletcher2000}
R.~Fletcher, \emph{{Practical Methods of Optimization}}.\hskip 1em plus 0.5em
  minus 0.4em\relax John Wiley \& Sons, 2000.

\bibitem{Nocedal2006}
J.~Nocedal and S.~J. Wright, \emph{{Numerical optimization}}, 2006.

\bibitem{Cotter2009}
S.~L. Cotter, M.~Dashti, J.~C. Robinson, and A.~M. Stuart, ``Bayesian inverse
  problems for functions and applications to fluid mechanics,'' \emph{Inverse
  Problems}, vol.~25, no.~11, 2009.

\bibitem{harris2020array}
\BIBentryALTinterwordspacing
C.~R. Harris, K.~J. Millman, S.~J. van~der Walt, R.~Gommers, P.~Virtanen,
  D.~Cournapeau, E.~Wieser, J.~Taylor, S.~Berg, N.~J. Smith, R.~Kern, M.~Picus,
  S.~Hoyer, M.~H. van Kerkwijk, M.~Brett, A.~Haldane, J.~F. del R{\'{i}}o,
  M.~Wiebe, P.~Peterson, P.~G{\'{e}}rard-Marchant, K.~Sheppard, T.~Reddy,
  W.~Weckesser, H.~Abbasi, C.~Gohlke, and T.~E. Oliphant, ``Array programming
  with {NumPy},'' \emph{Nature}, vol. 585, no. 7825, pp. 357--362, Sep. 2020.
  [Online]. Available: \url{https://doi.org/10.1038/s41586-020-2649-2}
\BIBentrySTDinterwordspacing

\bibitem{2020SciPy-NMeth}
P.~Virtanen \emph{et~al.}, ``{{SciPy} 1.0: Fundamental Algorithms for
  Scientific Computing in Python},'' \emph{Nature Methods}, vol.~17, pp.
  261--272, 2020.

\bibitem{Cheng2008}
\BIBentryALTinterwordspacing
N.-S. Cheng, ``{Formula for the Viscosity of a Glycerol--Water Mixture},''
  \emph{Industrial \& Engineering Chemistry Research}, vol.~47, no.~9, pp.
  3285--3288, may 2008. [Online]. Available:
  \url{https://doi.org/10.1021/ie071349z}
\BIBentrySTDinterwordspacing

\bibitem{Volk2018}
\BIBentryALTinterwordspacing
A.~Volk and C.~J. K{\"{a}}hler, ``{Density model for aqueous glycerol
  solutions},'' \emph{Experiments in Fluids}, vol.~59, no.~5, p.~75, 2018.
  [Online]. Available: \url{https://doi.org/10.1007/s00348-018-2527-y}
\BIBentrySTDinterwordspacing

\bibitem{Bouillot2018}
P.~Bouillot, B.~M. Delattre, O.~Brina, R.~Ouared, M.~Farhat, C.~Chnafa, D.~A.
  Steinman, K.~O. Lovblad, V.~M. Pereira, and M.~I. Vargas, ``{3D phase
  contrast MRI: Partial volume correction for robust blood flow quantification
  in small intracranial vessels},'' \emph{Magnetic Resonance in Medicine},
  vol.~79, no.~1, pp. 129--140, 2018.

\bibitem{Saito2020}
K.~Saito, S.~Abe, M.~Kumamoto, Y.~Uchihara, A.~Tanaka, K.~Sugie, M.~Ihara,
  M.~Koga, and H.~Yamagami, ``{Blood Flow Visualization and Wall Shear Stress
  Measurement of Carotid Arteries Using Vascular Vector Flow Mapping},''
  \emph{Ultrasound in Medicine and Biology}, vol.~46, no.~10, pp. 2692--2699,
  2020.

\bibitem{Gladden2017}
L.~F. Gladden and A.~J. Sederman, ``{Magnetic Resonance Imaging and Velocity
  Mapping in Chemical Engineering Applications},'' \emph{Annual Review of
  Chemical and Biomolecular Engineering}, vol.~8, no.~1, pp. 227--247, 2017.

\bibitem{Deshmane2012}
\BIBentryALTinterwordspacing
A.~Deshmane, V.~Gulani, M.~A. Griswold, and N.~Seiberlich, ``{Parallel MR
  imaging},'' \emph{Journal of Magnetic Resonance Imaging}, vol.~36, no.~1, pp.
  55--72, 2012. [Online]. Available:
  \url{https://onlinelibrary.wiley.com/doi/abs/10.1002/jmri.23639}
\BIBentrySTDinterwordspacing

\bibitem{Pruessmann1999}
\BIBentryALTinterwordspacing
K.~P. Pruessmann, M.~Weiger, M.~B. Scheidegger, and P.~Boesiger, ``{SENSE:
  Sensitivity encoding for fast MRI},'' \emph{Magnetic Resonance in Medicine},
  vol.~42, no.~5, pp. 952--962, 1999. [Online]. Available:
  \url{https://onlinelibrary.wiley.com/doi/abs/10.1002/%28SICI%291522-2594%28199911%2942%3A5%3C952%3A%3AAID-MRM16%3E3.0.CO%3B2-S}
\BIBentrySTDinterwordspacing

\bibitem{stein1970}
\BIBentryALTinterwordspacing
E.~M. Stein, \emph{{Singular {Integrals} and {Differentiability} {Properties}
  of {Functions} ({PMS}-30)}}.\hskip 1em plus 0.5em minus 0.4em\relax Princeton
  University Press, 1970. [Online]. Available:
  \url{http://www.jstor.org/stable/j.ctt1bpmb07}
\BIBentrySTDinterwordspacing

\bibitem{espirit2014}
\BIBentryALTinterwordspacing
M.~Uecker, P.~Lai, M.~J. Murphy, P.~Virtue, M.~Elad, J.~M. Pauly, S.~S.
  Vasanawala, and M.~Lustig, ``Espirit—an eigenvalue approach to
  autocalibrating parallel mri: Where sense meets grappa,'' \emph{Magnetic
  Resonance in Medicine}, vol.~71, no.~3, pp. 990--1001, 2014. [Online].
  Available: \url{https://onlinelibrary.wiley.com/doi/abs/10.1002/mrm.24751}
\BIBentrySTDinterwordspacing

\bibitem{Sun2017}
\BIBentryALTinterwordspacing
A.~Sun, B.~Zhao, K.~Ma, Z.~Zhou, L.~He, R.~Li, and C.~Yuan, ``Accelerated phase
  contrast flow imaging with direct complex difference reconstruction,''
  \emph{Magnetic Resonance in Medicine}, vol.~77, no.~3, pp. 1036--1048, 2017.
  [Online]. Available:
  \url{https://onlinelibrary.wiley.com/doi/abs/10.1002/mrm.26184}
\BIBentrySTDinterwordspacing

\end{thebibliography}

\end{document}